\documentclass[10pt,twocolumn,letterpaper]{article}

\usepackage{cvpr}              

\usepackage[dvipsnames]{xcolor}

\usepackage{graphicx}
\usepackage{amsmath}
\usepackage{amssymb}
\usepackage{booktabs}
\usepackage{xcolor}
\usepackage{multirow}
\usepackage[percent]{overpic}

\usepackage{array}
\newcolumntype{P}[1]{>{\centering\arraybackslash}p{#1}}

\usepackage{bm}
\usepackage{bbm}

\usepackage{multicol}
\usepackage{multirow}

\definecolor{cvprblue}{rgb}{0.21,0.49,0.74}
\usepackage[pagebackref,breaklinks,colorlinks,citecolor=cvprblue]{hyperref}

\usepackage[capitalize]{cleveref}
\crefname{section}{Sec.}{Secs.}
\Crefname{section}{Section}{Sections}
\Crefname{table}{Table}{Tables}
\crefname{table}{Tab.}{Tabs.}

\def\paperID{966} 
\def\confName{CVPR}
\def\confYear{2024}

\title{Relightful Harmonization: Lighting-aware Portrait Background Replacement }
\author{Mengwei Ren$^1{^2}{^*}$
\quad
Wei Xiong$^2$
\quad
Jae Shin Yoon$^2$
\quad
Zhixin Shu$^2$ \\
Jianming Zhang$^2$
\quad
HyunJoon Jung$^2$
\quad
Guido Gerig$^1$
\quad
He Zhang$^2$ \\ [0.2em]
$^1$New York University \qquad \qquad $^2$Adobe
}

\usepackage{footnote}
\usepackage[symbol]{footmisc}
\begin{document}

\twocolumn[{%
\renewcommand\twocolumn[1][]{#1}
\maketitle
\footnotetext[1]{Work done during an internship at Adobe.}
\renewcommand*{\thefootnote}{\arabic{footnote}}
\setcounter{footnote}{0}
\vspace{-8mm}
\begin{center}
    \centering
    \includegraphics[width=0.99\textwidth]{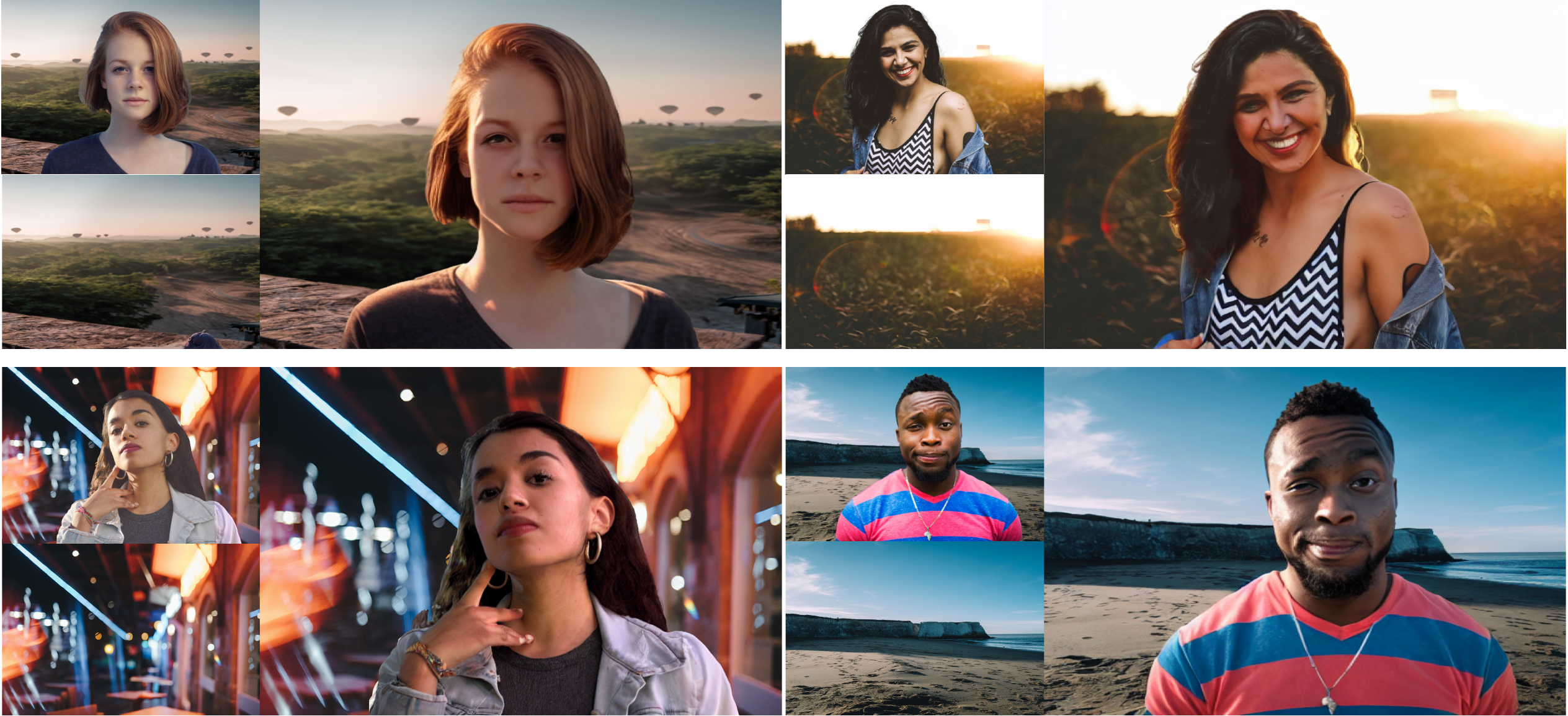}
    \captionof{figure}{Relightful Harmonization on four real-world images. Each set shows a direct composition (\textit{upper left}) of the foreground subject onto a new backgound (\textit{lower left}), and our harmonized result (\textit{right}) that accounts for both \textbf{lighting} and \textbf{color}.} 
    \label{fig:teaser}
\end{center}%
}]

\footnotetext[1]{Work done during an internship at Adobe.}
\renewcommand*{\thefootnote}{\fnsymbol{footnote}}
\setcounter{footnote}{0}
\begin{abstract}
Portrait harmonization aims to composite a subject into a new background, adjusting its lighting and color to ensure harmony with the background scene. 
Existing harmonization techniques often only focus on adjusting the global color and brightness of the foreground and ignore crucial illumination cues from the background such as apparent lighting direction, leading to unrealistic compositions.
We introduce Relightful Harmonization, a lighting-aware diffusion model designed to seamlessly harmonize sophisticated lighting effect for the foreground portrait using any background image.
Our approach unfolds in three stages. First, we introduce a lighting representation module that allows our diffusion model to encode lighting information from target image background.
Second, we introduce an alignment network that aligns lighting features learned from image background with lighting features learned from panorama environment maps, which is a complete representation for scene illumination.
Last, to further boost the photorealism of the proposed method, we introduce a novel data simulation pipeline that generates synthetic training pairs from a diverse range of natural images, which are used to refine the model.
Our method outperforms existing benchmarks in visual fidelity and lighting coherence, showing superior generalization in real-world testing scenarios, highlighting its versatility and practicality.

\end{abstract}

\vspace{-3mm}
\section{Introduction}
\label{sec:intro}
Portrait harmonization~\cite{valanarasu2022interactive,wang2023harmonized} stands as a crucial element in both photography and creative image editing, seeking to seamlessly composite a subject into a new background while maintaining realism and aesthetic uniformity in terms of lighting and color. The process initiates with the segmentation of the subject from its original image, followed by the composition into a new background. To enhance visual consistency, the subsequent step entails meticulous adjustments to the foreground, aligning it with the new background—considering factors such as color, brightness, saturation, and lighting conditions. The manual effort could be labor-intensive, particularly when dealing with intricate lighting scenarios in portraiture.

There are two principal sets of methods that could automatically adjust the foreground to match the background for human portraits: (1) image harmonization techniques, and (2) portrait relighting methods. 
Harmonization-based methods~\cite{cong2020dovenet,INR,PIH_wang2023semi,PCT_Guerreiro_2023_CVPR,Harmonizer,cong2022high,guo2021image,guo2021intrinsic,jiang2021ssh,liang2021spatial,xue2012understanding,tan2023deep,xing2022composite,lu2023painterly} aim to match the color statistics of the foreground with those of the background, by adjusting the foreground color tone, contrast, and illumination. Yet, they often overlook the lighting characteristics and leave the foreground illumination effects unchanged, such as the lighting direction and shadows, potentially resulting in an unnatural appearance when the background has distinct lighting conditions. For instance, compositing a person photographed under a top-down light into a sunset scene might make the composite non-realistic to the human eye. 
On the other hand, recent work on portrait relighting~\cite{nestmeyer2020learning,pandey2021total,sun2019single,zhou2019deep,zhang2021neural,nvpr, wang2020single, yeh2022learning, Mei_2023_CVPR} are designed for matching the lighting of the subject towards a new environment by using the paired training data acquired with the light stage system~\cite{debevec2000acquiring}. 
Nevertheless, current relighting methods often require dynamic range (HDR) panorama environment maps~\cite{pandey2021total} during training and inference, which are not always feasible to acquire, especially in casual photography settings. 
In this work, we explore the possibility of generating realistic and lighting-aware composition images in a straightforward harmonization set up. Given a foreground image (with its corresponding alpha mask) and an arbitrary background image, we propose a unified and end-to-end framework that encompasses both \emph{color} and \emph{lighting} harmonization. We approach the task through a conditional generative framework, leveraging a pretrained diffusion model~\cite{cong2022high,InstructPix2Pix}, and develop a three-stage training pipeline.

In the first phase, we conduct \textbf{Lighting-aware Diffusion Training} to integrate explicit lighting conditioning into a pretrained diffusion model. This involves a lighting representation learning module that derives lighting conditions from a selected background image. The resulting lighting information is then integrated into the diffusion UNet backbone to guide the generative process. The training is performed on a pairwise relighting-specialized light stage dataset to effectively capture the lighting dynamics.

Given the challenge of accurately inferring environmental lighting from a single background image, which is inherently an ill-posed problem, we employ paired environment maps to augment the physical plausibility of our background-derived lighting representation. This is achieved through a second stage of \textbf{Lighting Representation Alignment}, designed to align the lighting representation extracted from background images with that learned from their corresponding panorama environment maps. 

Finally, we perform \textbf{Finetuning for Photorealism} on an expanded dataset to improve the photorealism of the harmonization. We propose a novel data synthesis pipeline using our initially trained model to create additional data from natural images.
Notably, once trained, our pipeline \emph{does not} rely on any external environment maps, which greatly empowers the proposed framework for flexible background replacement and portrait harmonization.

Our contributions are threefold. 
(1) We enable the lighting effects to be encoded in a pretrained image-conditioned latent diffusion model by incorporating a spatial lighting feature extraction and conditioning module to the diffusion backbone. The background-extracted lighting representation is further aligned with the feature extracted from panorama environment map to ensure better physical plausibility.
(2) We use our model as a data augmenter and propose a novel data simulation pipeline to synthesize training pairs from natural images. The model is then refined with the enlarged dataset to further boost the photorealism of the results. 
(3) Compared with existing harmonization and relighting methods, our pipeline demonstrates improvements of the harmonized results in both lighting coherence visual fidelity, providing a versatile solution for real-world portrait harmonization in a variety of settings.

\section{Related Works}

\noindent \textbf{Image Harmonization} aims to rectify color, contrast, and style differences between foreground and background to ensure natural and consistent composition. In deep learning, this task is approached as an end-to-end image-to-image translation problem~\cite{zhu2015learning,tsai2017deep,guo2021intrinsic,jiang2021ssh,cong2020dovenet,guo2021image,cong2022high,PIH_wang2023semi,INR,PCT_Guerreiro_2023_CVPR,tan2023deep,xing2022composite,lu2023painterly}, where the network is trained to predict a harmonized image from the input composite. Pixel-aligned datasets are created by altering foreground color in real images with pre-designed~\cite{cong2020dovenet} or learned~\cite{SycoNet2023} augmentations.  
Yet, existing harmonization methods primarily focus on global color adjustment, overlooking the subtle but important discrepancies between foreground and background lighting, i.e., direction, intensity, and shadow. This can lead to a harmonized image that, despite matching colors, still appears unnatural due to mismatched lighting conditions. Therefore, we postulate that enhancing the lighting-awareness of harmonization models is a vital yet underexplored area for natural and realistic composition.\\

\begin{figure*}[ht]
\centering
\vspace{-8mm}
\includegraphics[width=0.9\linewidth]{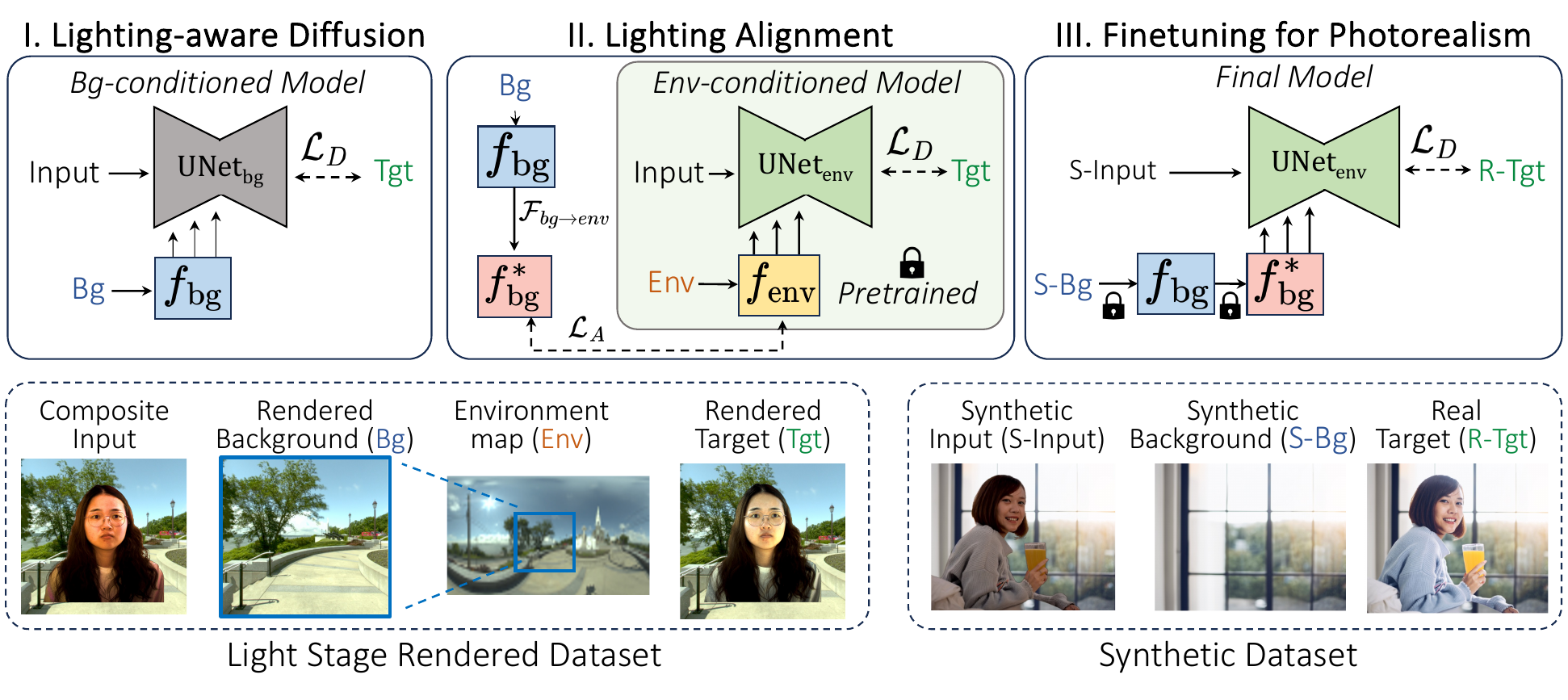}
\vspace{-4mm}
    \caption{The Pipeline of Relightful Harmonization. In Stage I, a lighting representation module is integrated into the diffusion model, conditioning the generation on lighting information encoded from the background image, trained with a light stage dataset for relighting (\textit{lower left}). Stage II aligns lighting features derived from the background with the environment map for enhanced physical accuracy. Finally, Stage III refines the model on a real image dataset (\textit{lower right}) obtained via a novel data simulation pipeline. }
    \label{fig:overview}
\vspace{-4mm}
\end{figure*}
\noindent \textbf{Portrait Relighting} Recent advancements in portrait relighting has been driven by deep learning methods~\cite{nestmeyer2020learning,pandey2021total,sun2019single,zhou2019deep,zhang2021neural,nvpr, wang2020single, yeh2022learning, Mei_2023_CVPR,ponglertnapakorn2023difareli,mei2024holo}. They leverage supervised training with the paired training data acquired with the light stage system~\cite{debevec2000acquiring}. These methods require a target HDR enviroment map as the external input source, and typically involve the intermediate prediction of surface normals, albedo, and/or a set of diffuse and specular maps with ground truth supervision.
However, reliance on HDR maps for background replacement and harmonization tasks poses significant limitations on their applicability in everyday scenarios where HDR maps cannot be easily captured alongside~\cite{pandey2021total}. In our framework, we do not require any external input sources except a single target background image. 
Furthermore, many current relighting systems employ multistage frameworks and/or heavily rely on external packages~\cite{feng2021learning}. The accuracy and performance of these systems are consequently contingent on the precision of each individual stage, making the overall process complicated and prone to errors propagated through these intermediate steps.
Additionally, the datasets commonly employed for training are rendered from limited light stage illumination acquisition, which means the target images utilized during the training phase are not captured in real-world conditions but are rendered composites. Therefore, its generalization on unseen images, as well as the photorealism when applied to arbitrary background replacement tasks remain unclear.

\noindent \textbf{Diffusion Models}~\cite{sohl2015deep,ho2020denoising,ddim,dhariwal2021diffusion} have significantly advanced the image and video synthesis quality~\cite{ramesh2022hierarchical,rombach2022high,ho2021classifier,kawar2022enhancing,ho2022video,kawar2022jpeg}. 
Image-conditioned diffusion models ~\cite{ControlNet,wang2022pretraining,saharia2022palette} typically take an image as additional input to perform an image-conditioned generation such as image enhancement~\cite{li2022srdiff,
ddrm,ren2023multiscale,kawar2022jpeg}, harmonization~\cite{chen2023zeroshot,lu2023tf,painterly} and translation~\cite{ControlNet,sasaki2021unit,zhao2022egsde,wang2022pretraining,kwon2022diffusion,nie2023blessing}, typically trained on task-specific pairwise data.
Recently, the application of pretrained text-to-image diffusion models~\cite{nichol2021glide,rombach2022high, saharia2022photorealistic} has been extended to image editing tasks~\cite{meng2021sdedit,hertz2022prompt,couairon2022diffedit,InstructPix2Pix,mokady2022null,wallace2022edict,brack2023sega,tumanyan2023plug,miyake2023negative,zhang2023sine,wu2023latent,huberman2023edit,cao2023masactrl,alaluf2023cross,fu2023guiding,song2023objectstitch,song2024imprint,han2024proxedit}. These models leverage text-image correlations to perform context modifications in image editing, such as `adding a sunset' with InstructPix2Pix~\cite{InstructPix2Pix}, which is loosely connected to our lighting-aware set up. However, text-based editing does not incorporate spatial information, thus lacking finegrained control to the model. Instead, we propose to use a spatial lighting representation as the new `instruction' that guide the diffusion model to perform lighting-aware editing.

\section{Method}
We aim to develop a conditional diffusion model that processes a composite image (along with its alpha mask) as the input, conditioned on the target background, and produce \textit{color} and \textit{lighting} harmonized output. To do so, we develop a three-stage training strategy presented in Fig.~\ref{fig:overview}.  

\noindent\textbf{Stage I: Lighting-aware Diffusion Training:} We build our model on a pretrained diffusion model~\cite{InstructPix2Pix} and enable its lighting awareness by attaching a lighting representation learning branch to encode lighting information from the background image, which is then injected into the UNet backbone as illustrated in Fig.~\ref{fig:overview}-I. The training is conducted with relighting-specialized light stage rendered dataset, as shown on the bottom left of Fig.~\ref{fig:overview}-I.

\noindent\textbf{Stage II: Lighting Representation Alignment:}
As one of our of goals is to enable lighting-aware portrait harmonization \textit{without} relying on environment maps during inference, we propose a representation alignment step (Fig.~\ref{fig:overview}-II) to adapt the lighting representation extracted from a background image towards the learned representation from its environment map. We assume the aligned representation is more robust and physically plausible.

\noindent\textbf{Stage III: Finetuning for Photorealism:}
In the third stage, we finetune our model using high-quality pixel-aligned training pairs from natural images, where these paired datasets are generated via a novel data synthesis pipeline (Fig.~\ref{fig:overview}-III) using the stage 2 model as a data augmenter.

\subsection{Lighting-aware Diffusion Training} 
\label{sec:stage1}
Our lighting-aware diffusion model learns to generate a harmonized image given the composite input, conditioned on a lighting representation extracted from the target background image. 
Adhering to established practice for training relighting models~\cite{pandey2021total,zhang2020portrait,sun2019single}, we assume access to the light stage rendered training dataset. A training tuple includes the input image (alongside its alpha mask), the target background, the target environment map, as well as the target image.
An example of training data is shown on the bottom left of Fig.~\ref{fig:overview}. 

Formally, we represent a rendered image sample as ${x_i^a}$, indicating a portrait image of subject $i$ illuminated under the lighting condition $a$. The corresponding environment map is denoted as $z_i^a$, and the background image ${y_i^a}$ for ${x_i^a}$ is generated by projecting the HDR map with a specified field of view and resolution. Subject masks $m_i$ are obtained using methods described in ~\cite{Zhang_2021_WACV,Yu_2021_CVPR}.
\noindent\\\textbf{Lighting Conditioning:} We further modify the diffusion backbone to incorporate explicit lighting conditions. As depicted in Fig.~\ref{fig:overview}-I, a lighting conditioning branch is integrated atop the UNet backbone, injecting a lighting representation $f$ encoded from the target background image by a CNN $\mathcal{F}$, at multiple scales within the UNet. The conditioning mechanism is designed in a similar fashion as ~\cite{ControlNet} where conditional feature maps are added to the UNet features at respective resolution within the encoder. 

With the light stage dataset, we postulate that the lighting representation $f$ can be learned from the pairwise training. Specifically, a training tuple from the same subject $i$ is sampled as $(x^a, m, y^b, x^b)$. Noise is progressively added to the target image $x^b$ until time step $t$, resulting in a noisy image $x_t^b$. The UNet, denoted as $\mathcal{U}_{bg}$, is conditioned on the background-extracted lighting feature $F_{bg}(y^b)$, and is trained to predict the noise $\epsilon$, with the following objective: 
\begin{align}
\mathcal{L}_{\text{D}} = \mathbb{E}_{x^a, y^b, x^b_0, t,\epsilon} \Big[ \left\| \epsilon - \mathcal{U}_{bg}(x^b_t, t, x^a, F_{bg}(y^b)) \right\|^2_2 \Big]
\label{eq:L_dpm}
\end{align}
where $\epsilon \sim \mathcal{N}(0,1)$. 
At this stage, we initilize the weights of the UNet backbone from~\cite{InstructPix2Pix}, and jointly train \textit{both} UNet and the conditioning branch. 

\subsection{Lighting Representation Alignment}
\label{sec:stage2}
Given that a background image is a partial projection of the environment map which encapsulates panoramic lighting information (see example on bottom left of Fig.~\ref{fig:overview}), we suspect that the lighting cues learned from an environment map will be inherently more comprehensive than those from $f_{bg}$, implying that an environment-conditioned model could potentially offer more physically plausible relighting under the same training scenario. This is further empirically verified in our ablation detailed in Sec.~\ref{sec:ablation}. 
However, in real-world photography, environment maps are usually not co-acquired which poses practical challenges, limiting the applicability of environment map dependent models. To circumvent this, we align the lighting representation extracted from a background image with features derived from its ground truth environment map, ultimately enabling effective portrait harmonization with just a single background image.

As shown in Fig.~\ref{fig:overview} II., we first pretrain an environment map conditioned harmonization model (in green box) to generate a ground truth environment map-derived lighting representation. The model architecture is identical to the background-based model in Fig.~\ref{fig:overview} I, while substituting the Stage I input condition from the background image to its corresponding environment map. 
It is trained with the same light stage dataset, under a denoising loss analogous to Eq.\ref{eq:L_dpm}, where the background lighting feature $f_{bg}$ is replaced with  an environment map-derived feature $\mathcal{F}_{\text{env}}(z^b)$.

Then, we freeze the environment-conditioned model and introduce an alignment network $\mathcal{F}_{bg\rightarrow env}$ that calibrates the background-extracted lighting representation to align with its environment map equivalent. 
We formulate such a process as an inverse problem that can be learned with a network $\mathcal{F}_\text{bg$\rightarrow$env}$, under a supervised loss. For a training tuple $(x^a, m, y^b, z^b, x^b)$, the alignment network takes $f_{\text{bg}}=\mathcal{F}_{\text{bg}}(y^b)$ as input, and maps it with the alignment network to $f^*_\text{bg}$.
The environment extracted feature $f_{\text{env}}=\mathcal{F}_\text{env}(z^b)$ is utilized as the ground truth, and we use a $L_1$ objective:
\begin{align}
f^* &= F_{\text{bg} \to \text{env}}(F_{bg}(y^b)),\nonumber\\
\mathcal{L_\text{A}} &= \mathbb{E}_{y^b, z^b} \Big[ \left\| F_{\text{env}}(z^b) - f^*_\text{bg} \right\|_1\Big].
\end{align}
During this phase, we update only $\mathcal{F}{\text{bg}\rightarrow \text{env}}$, while freezing the other networks.
We assume that this alignment enhances the background-derived lighting representation to more accurately encode the environmental lighting, which is empirically verified where the aligned feature maps reflect more global illumination information (Fig.~\ref{fig:feature}). Once trained, we integrate the aligned feature extraction and conditioning into $\mathcal{U}_\text{env}$, formulating our final model in Fig.~\ref{fig:overview}-III.

\subsection{Finetuning for Photorealism} 
\label{sec:stage3}
The light stage dataset serves as a valuable resource for learning lighting representations, providing physically constrainted relighting pairs. However, it is essential to recognize that this dataset is \emph{compositional} by nature. It uses backgrounds projected from environment maps, combined with relit foreground subjects, to create composites that serve as ground truth for the diffusion model. 
However, these composites differ from real photographs, leading to potential concerns about the photorealism of the model's outputs.
Moreover, due to the cost of light stage data acquisition, it restricts the number and diversity of the subjects that can be collected. The diversity of the background images is also bounded by the scale of available environment maps during rendering (a few thousands). These limitations could affect the model's generalization ability to real images, and the capacity to produce realistic and varied lighting effects.
\begin{figure}[t]
\centering
\includegraphics[width=\linewidth]{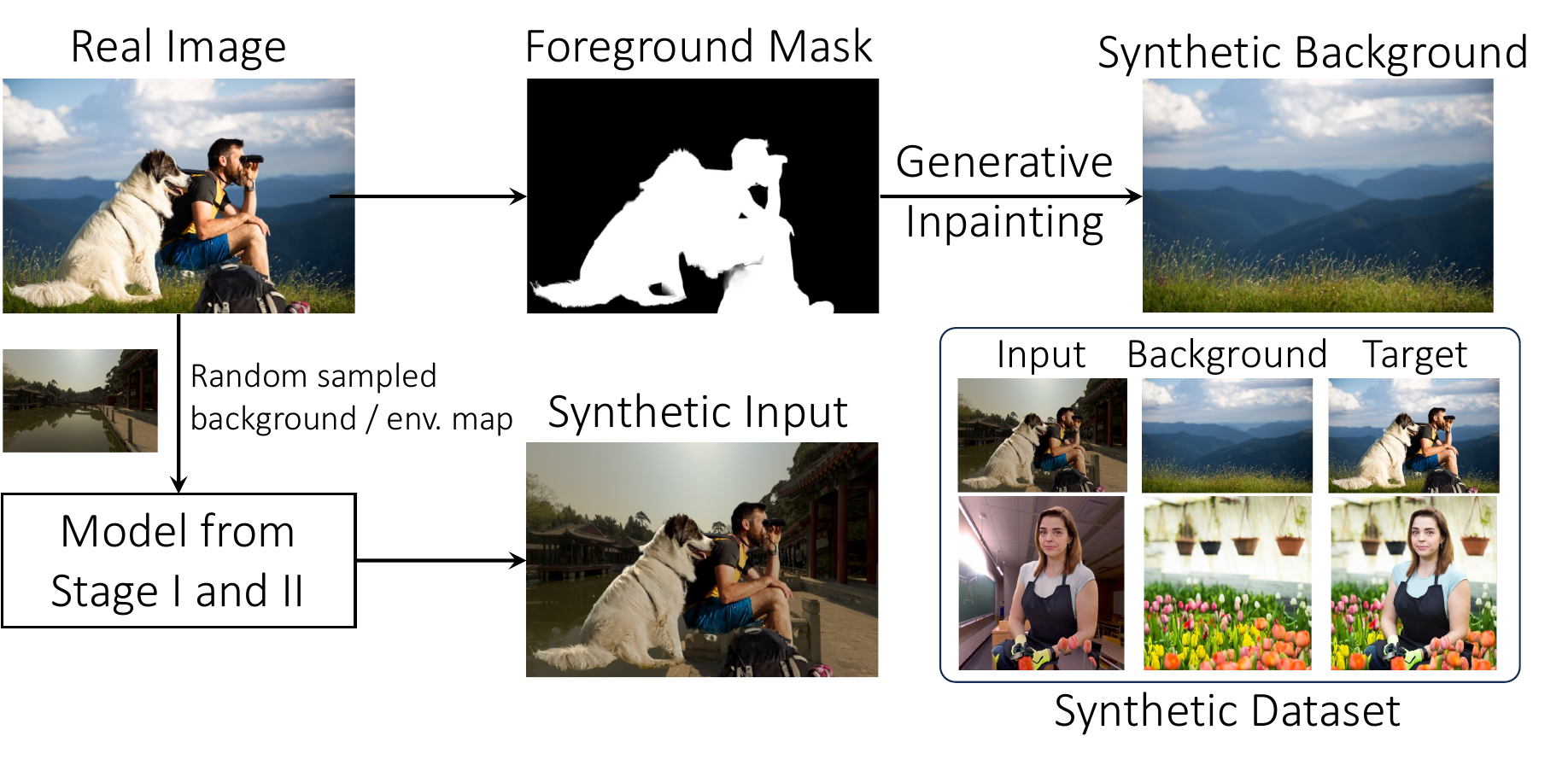}
\vspace{-5mm}
    \caption{Data synthesis pipeline. Given a real image, and inpaint the subject region, we obtain a synthetic background. The foreground lighting is then altered with our model trained in Stage I/II, to create an input image with distinct lighting. Two example pairs  are shown on the lower right. }
    \label{fig:data}
\vspace{-5mm}
\end{figure}
Therefore, we propose a third stage (Fig.~\ref{fig:overview} III) that
finetunes our final model for improved photorealism.

We introduce a novel data synthesis pipeline that creates pairwise training pairs from natural images, to ensure that the ground truth for finetuning the diffusion model remains \textit{real} images. As depicted in Fig.~\ref{fig:data}, the process starts with a portrait photograph, from which we extract the foreground mask~\cite{Zhang_2021_WACV,Yu_2021_CVPR}. We then inpaint the foreground region~\cite{xie2023smartbrush} with text guidance `clear background' to create a clean background image for the real image, which can serve as the condition input for the training. Next, the lighting of the foreground subject(s) is altered by running our trained model from stage I/II with a randomly chosen background image or environment map as the condition. This produces a synthetic input image with distinct foreground lighting and color compared to the target image.
Two sets of generated training tuples are displayed at bottom right of Fig.~\ref{fig:data}.
Once we obtain a sufficient number of synthetic data, we combine the original light stage dataset with the synthetic data to refine our model. During this stage, we freeze the lighting representation extraction and conditioning branch and only finetune the UNet backbone to refine the synthesis quality while maintaining the learned lighting plausibility. 
Once trained, our final model in Fig.~\ref{fig:overview} III is used to perform portrait harmonization given arbitrary background images, eliminating the need for environment maps.

\section{Experiments}
\subsection{Setup and Metrics}
Three testing scenarios are created for evaluation: (1) 500 Light stage rendered test pairs to evaluate the lighting accuracy; (2) 200 natural image test pairs, synthetically created from real images using our data synthesis pipeline, to assess the lighting realism; and most importantly, (3) Real-world portraits combined with arbitrary backgrounds, examining the model's generalizability and adaptability in real-life scenarios. For (1) and (2), we also quantify the results with common metrics MSE, SSIM, PSNR and LPIPS~\cite{zhang2018perceptual}.
To benchmark, we compare Relightful Harmonization with both established harmonization methods \texttt{INR}~\cite{INR}, \texttt{PCT}~\cite{PCT_Guerreiro_2023_CVPR}, \texttt{Harmonizer}~\cite{Harmonizer} and \texttt{PIH}~\cite{PIH_wang2023semi}, and relighting method \texttt{TR}~\cite{pandey2021total}. We also construct a relighting baseline with a \texttt{transformer} architecture and trained it with light stage data. More details are provided in the appendix. Note that relighting methods are applied only on the light stage test set as they are not applicable without HDR maps. 

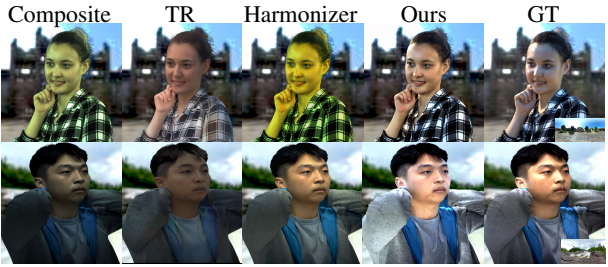
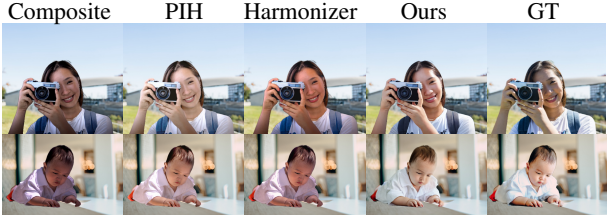
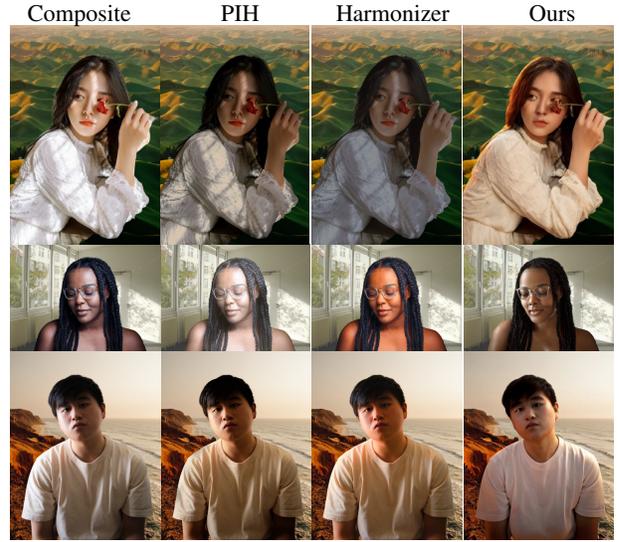
\begin{figure*}[ht]
  \centering
\vspace{-4mm}
  \begin{minipage}{.46\textwidth}
    \centering
    \begin{subfigure}{\textwidth}
      \centering
         \begin{overpic}[width=\textwidth]{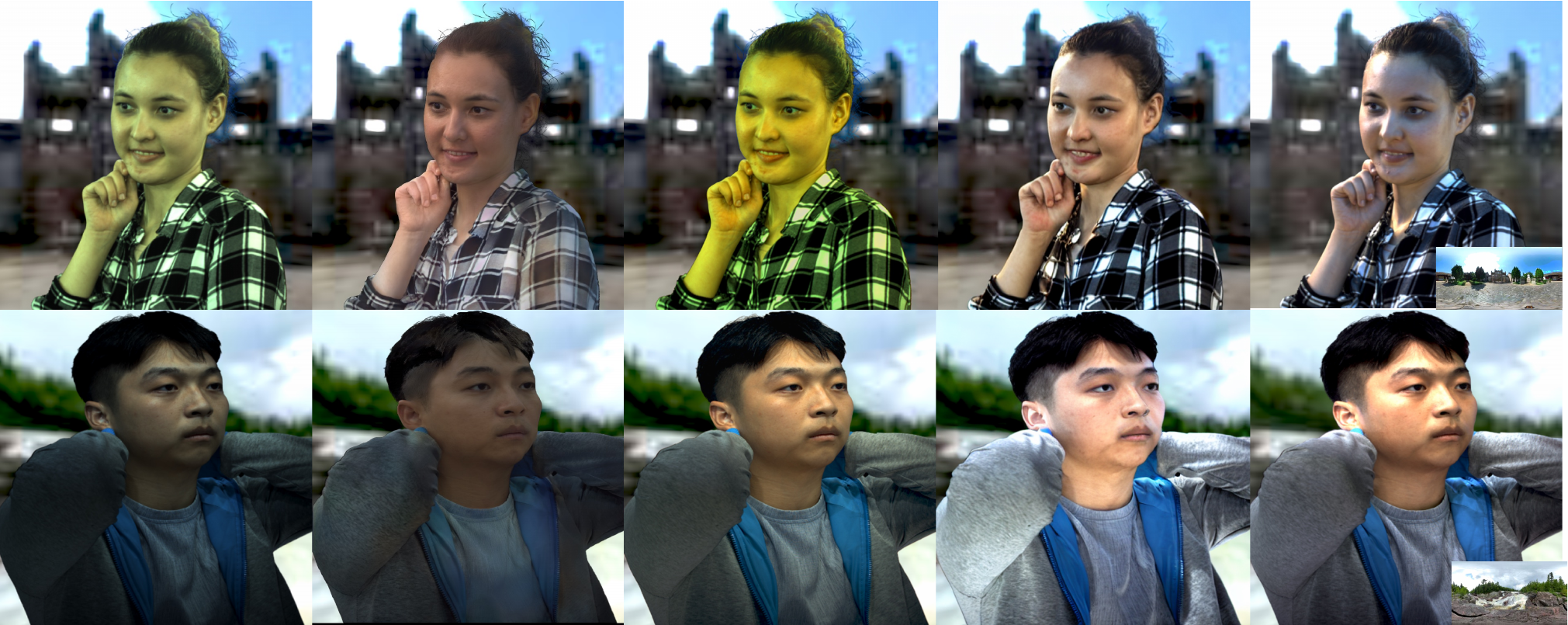}
    \put(1,40){\small Composite}
    \put(27,40){\small TR}
    \put(40,40){\small Harmonizer}
    \put(66,40){\small Ours}
    \put(87,40){\small GT}
    \end{overpic}
      \caption{Comparison on the light stage test set with ~\cite{pandey2021total} and ~\cite{Harmonizer}.}
      \label{fig:lightstage}
    \end{subfigure}
    \begin{subfigure}{\textwidth}
      \centering
        \vspace{12pt}
    \begin{overpic}[width=\textwidth]{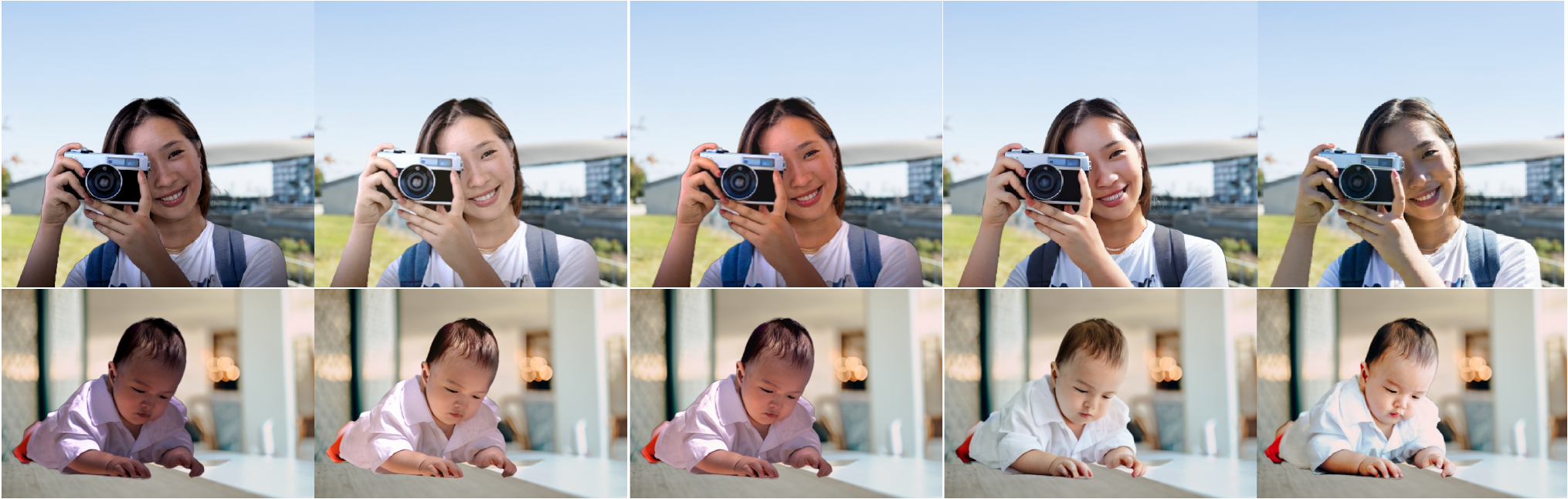}
    \put(1,32.5){\small Composite}
    \put(27,32.5){\small PIH}
    \put(40,32.5){\small Harmonizer}
    \put(66,32.5){\small Ours}
    \put(87,32.5){\small GT}
    \end{overpic}
      \caption{Comparison on the natural image synthetic test set with ~\cite{PIH_wang2023semi,Harmonizer}.}
      \label{fig:stock}
    \end{subfigure}
  \end{minipage}
  \hfill
  \begin{minipage}{0.46\textwidth}
  \begin{subfigure}{\textwidth}
    \centering
    \vspace{3pt}
    \begin{overpic}[width=\textwidth]{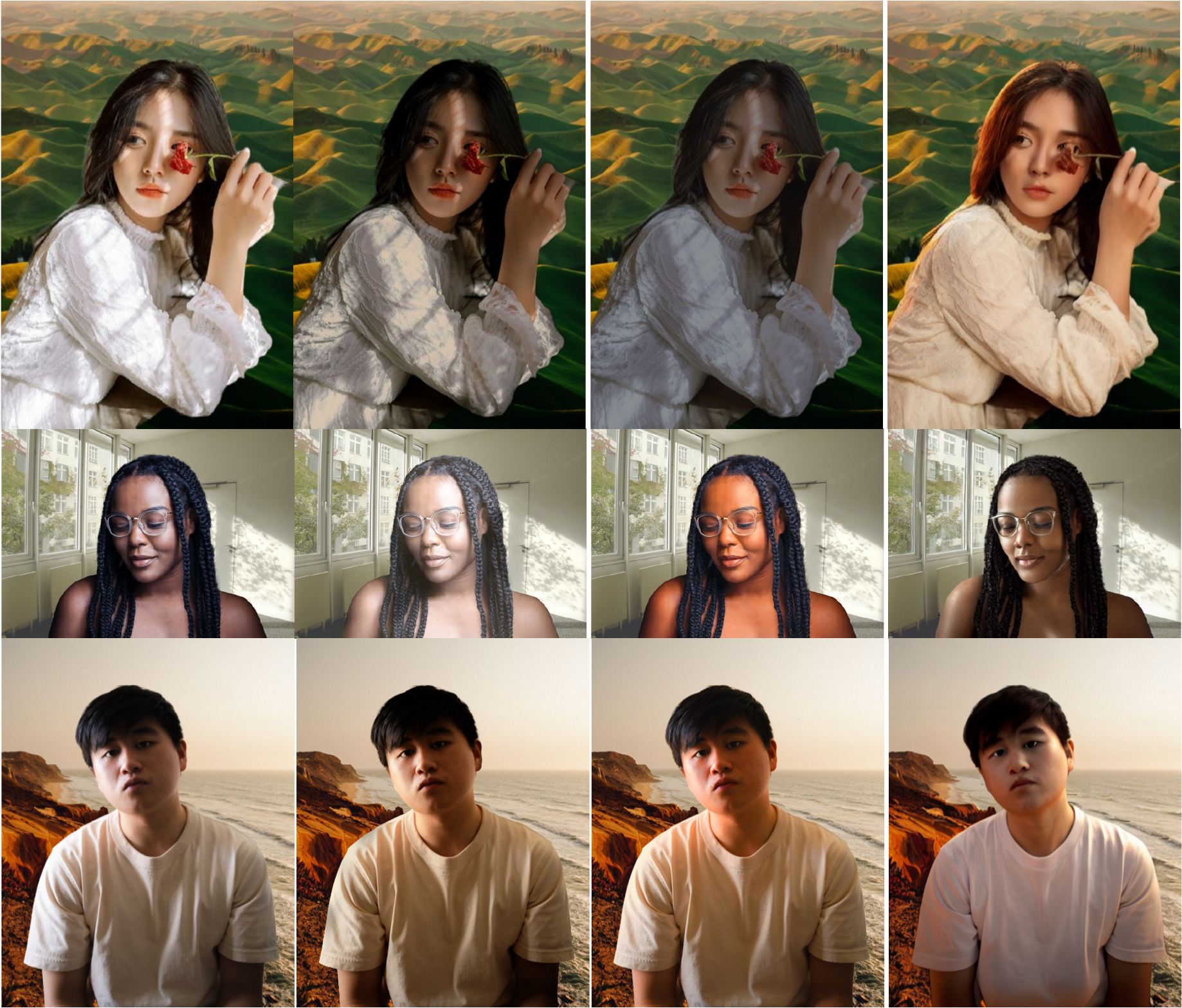}
    \put(3,86){\small Composite}
    \put(35,86){\small PIH}
    \put(54,86){\small Harmonizer}
    \put(86,86){\small Ours}
    \end{overpic}
    \caption{Comparison with~\cite{PIH_wang2023semi} and~\cite{Harmonizer} on real test images. Our method more effectively harmonizes incoherent foreground lighting and shadow.}
    \label{fig:benchmark-test}
\end{subfigure}
  \end{minipage}
\vspace{-3mm}
  \caption{Visual comparisons with benchmark methods. }
  \label{fig:benchmark_vis}
\end{figure*}

\begin{table*}[t]
\footnotesize
\centering
\caption{Quantitative results on both light stage test set and the natural image test set. The relighing methods including TR~\cite{pandey2021total} and the transformer baseline require HDR maps during the inference and are thus non-applicable on the natural image test set.}
\vspace{-4mm}
\label{table:benchmark}
\setlength{\tabcolsep}{4pt}
\begin{tabular}{lcccccccc}
\toprule
\multirow{2}{*}{Method} & \multicolumn{4}{c}{Light stage test set} & \multicolumn{4}{c}{Natural image test set} \\ 
\cmidrule(lr){2-5}\cmidrule(lr){6-9}
 & MSE $\downarrow$ & PSNR $\uparrow$ & SSIM $\uparrow$ & LPIPS $\downarrow$ & MSE $\downarrow$ & PSNR $\uparrow$ & SSIM $\uparrow$ & LPIPS $\downarrow$ \\ 
\midrule
TR~\cite{pandey2021total} & 0.044 (0.057) & 15.889 (4.318) & 0.757 (0.087) & 0.354 (0.092) & N/A & N/A & N/A & N/A \\
Transformer* & 0.026 (0.021) & 17.259 (3.715) & 0.742 (0.096) & 0.337 (0.095) & N/A & N/A & N/A & N/A \\
INR~\cite{INR} & 0.016 (0.014) & 19.147 (3.353) & 0.823 (0.081) & 0.327 (0.083) & 0.009 (0.005) & 21.566 (2.943) & 0.904 (0.038) & 0.113 (0.031) \\
Harmonizer~\cite{Harmonizer} & 0.015 (0.011) & 19.304 (2.980) & 0.822 (0.077) & 0.338 (0.087) & 0.010 (0.007) & 21.419 (3.506) & 0.905 (0.039) & 0.108 (0.032) \\
PCT~\cite{PCT_Guerreiro_2023_CVPR} & 0.020 (0.016) & 18.339 (3.454) & 0.808 (0.093) & 0.408 (0.082) & 0.014 (0.010) & 19.647 (3.279) & 0.898 (0.038) & 0.147 (0.039) \\
PIH~\cite{PIH_wang2023semi} & 0.018 (0.015) & 18.865 (3.087) & 0.807 (0.087) & 0.330 (0.089) & 0.010 (0.007) & 21.147 (3.097) & 0.901 (0.038) & 0.112 (0.033) \\
Ours & \textbf{0.012 (0.010)} & \textbf{20.527 (3.136)} & \textbf{0.848 (0.076)} & \textbf{0.159 (0.058)} & \textbf{0.005 (0.004)} & \textbf{23.562 (2.830)} & \textbf{0.913 (0.034)} & \textbf{0.097 (0.044)} \\
\bottomrule
\end{tabular}
\end{table*}

\subsection{Implementation Details}
Our model is implemented in PyTorch~\cite{paszke2019pytorch} using $8\times80$GB A100 at $512\times512$ resolutions, with 96 batch size. In stage I, we initialize UNet from the pretrained weights of the InstructPix2Pix~\cite{InstructPix2Pix} checkpoint. In the first and second stage, we use in total ~400$k$ training image pairs, rendered from a arbitrary combination of 100 unique light stage subjects, and 2908 HDR environment maps. We also randomly rotate the HDR maps and use various FoVs to increase the diversity of the background. In the third stage, we train the network with additional ~200$k$ pairs of images synthesized from natural images. We set learning rate to $5e{-5}$. 
More details are provided in the appendix.
\vspace{.5em}

\subsection{Benchmark Results}
 In Figs.~\ref{fig:lightstage} and~\ref{fig:stock}, we present visual comparisons of our method with selected benchmarks across on both test sets. Our approach performs better in adjusting both the foreground color and lighting, aligning more closely with the ground truth on the right. Quantitative results in Table~\ref{table:benchmark} further demonstrate the benefit of our methods.
Fig.~\ref{fig:benchmark-test} showcases test results on natural images. While harmonization methods adjust for color, the composition still lacks fidelity due to counterintuitive lighting on the composed image. For example, the strong cast shadow on the foreground in the first row, and opposite lighting directions between foreground/background in the second and third rows.
Full visual results with all methods are provided in appendix. 

\begin{table}[h]
\footnotesize
\setlength{\tabcolsep}{3pt}
\centering
\vspace{-1pt}
\caption{User preference. Each value represents  the fraction of times that raters preferred our results than the baseline method.}
\vspace{-3pt}
\label{table:user-study-transposed}
\begin{tabular}{lcccc}
\toprule
 Method & PIH~\cite{PIH_wang2023semi} & INR~\cite{INR} & PCT~\cite{PCT_Guerreiro_2023_CVPR} & Harmonizer~\cite{Harmonizer}\\
\midrule
Preference Rate & 0.713 & 0.702 & 0.845 & 0.639 \\
\bottomrule
\end{tabular}
\vspace{-4mm}
\end{table}
We further conducted a user study to verify the visual plausibility of our method on the real world testing set. Given pairwise comparison between our method and each baseline, we asked Amazon Mechanical Turk raters to select the better harmonized image from a given pair of results sample from 70 image. The results on 1750 ratings are collected, and we report the fraction of times that raters preferred our results over the baseline method in Table.~\ref{table:user-study-transposed}. 

 \begin{figure*}
\centering
\vspace{-5mm}
\begin{subfigure}[][][t]{0.96\textwidth}
    \includegraphics[width=\textwidth]{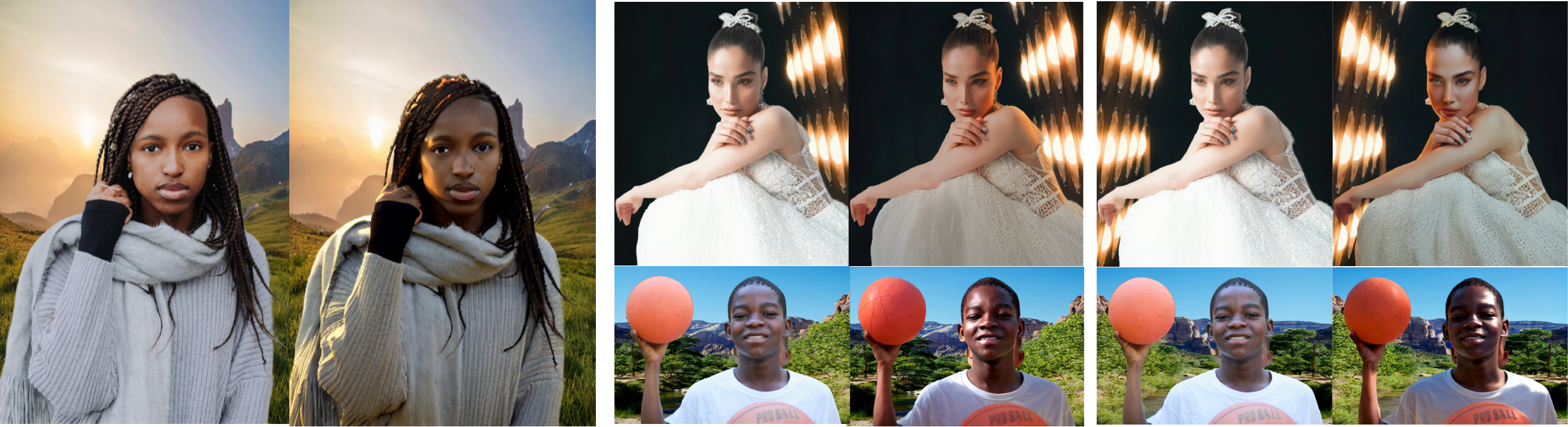}
    \caption{In cases where the target backgrounds offer clear lighting cues, our method generates visually convincing \textbf{lighting effects}. Additionally, upon flipping the background, we note consistent and appropriate adjustments to the lighting direction in the output.}
    \label{fig:flip}
\end{subfigure}
\unskip\ 
\begin{subfigure}[][][t]{0.97\textwidth}
    \includegraphics[width=\textwidth]{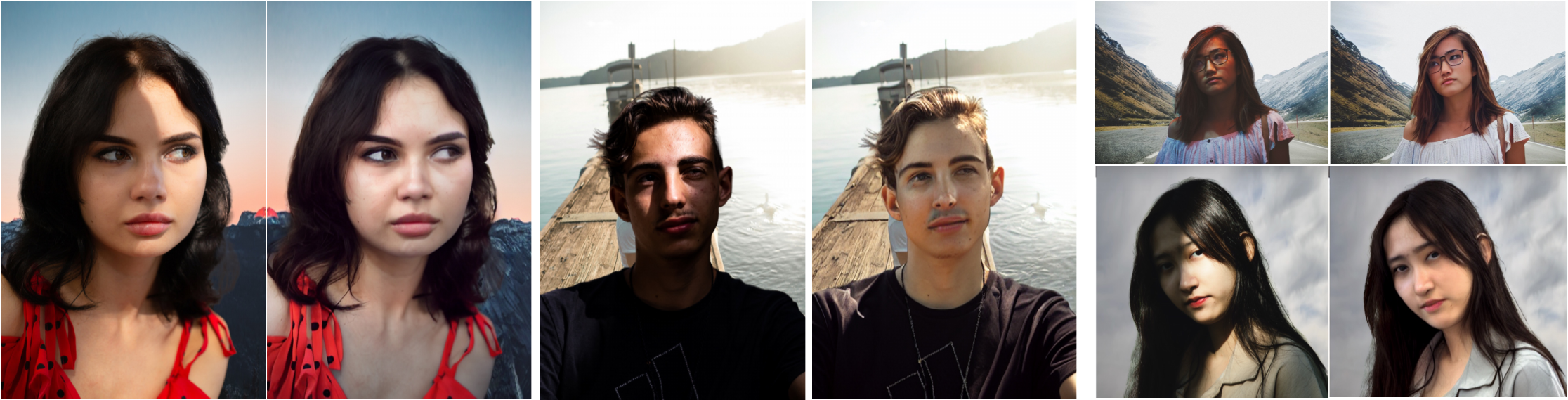}
    \caption{Our method effectively \textbf{neutralizes pronounced shadows} in the input while accommodating the ambient lighting of the background.}
    \label{fig:shadow-remove}
\end{subfigure}
\unskip\
\begin{subfigure}[][][t]{0.97\textwidth}
    \includegraphics[width=\textwidth]{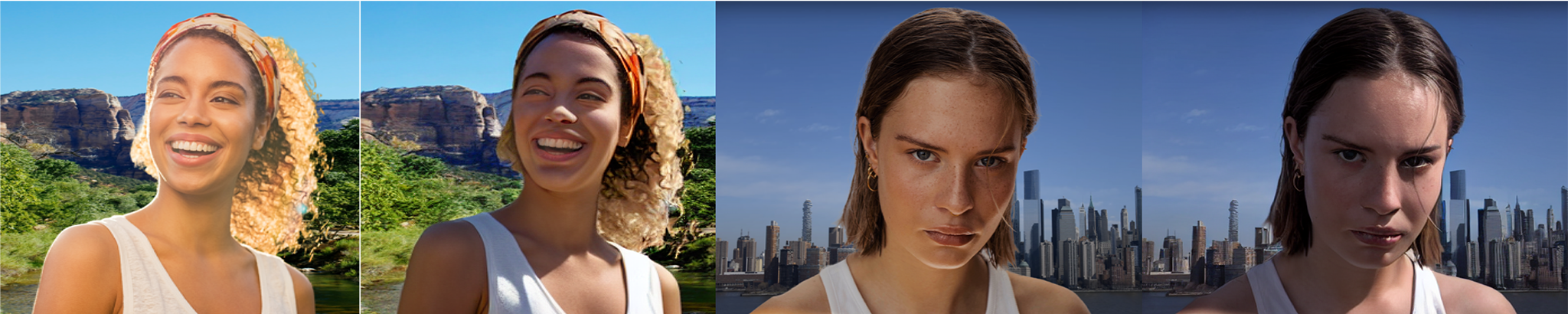}
    \caption{When applied to backgrounds with intense lighting conditions, e.g., with overhead sunlight, our method casts plausible \textbf{self-occlusion shadows}.}
    \label{fig:shadow-synth}
\end{subfigure}

\unskip\ 
\begin{subfigure}[][][t]{0.97\textwidth}
    \includegraphics[width=\textwidth]{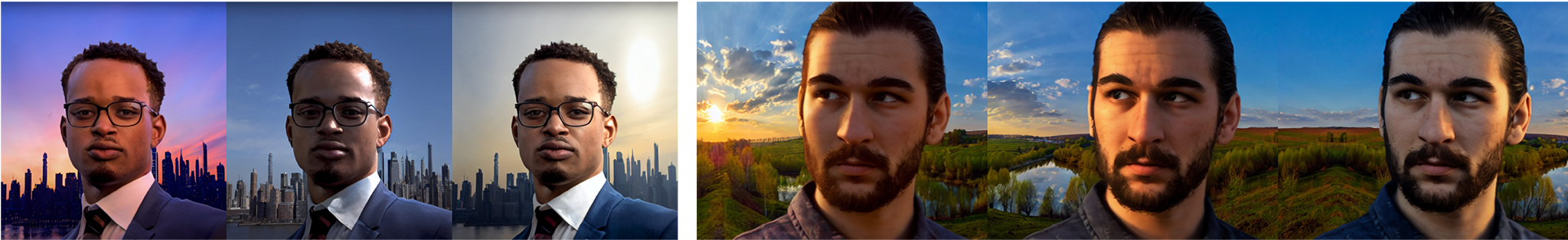}
    \caption{Our method consistently adjusts lighting when applied to moving backgrounds with {\textbf{temporally}}(left) and {\textbf{spatially}}(right) changing lighting directions.}
    \label{fig:spatiotemporal}
\end{subfigure}

\begin{subfigure}[][][t]{0.97\textwidth}
    \includegraphics[width=\textwidth]{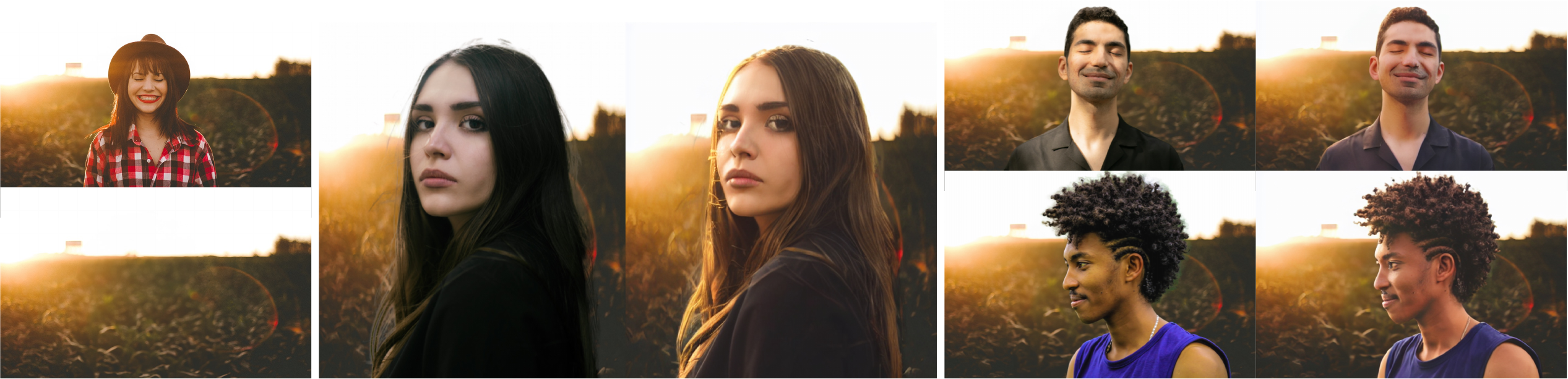}
    \caption{Our approach allows for reference-based harmonization tasks. This involves removing the subject from the reference image (\textit{upper left}) to create a background (\textit{lower left}) for composition. The harmonized results (\textit{right}) achieve lighting effects closely resembling those in the reference.}
    \label{fig:reference}
\end{subfigure}
\hfill
\caption{Real-world testing results under different scenarios to examine the lighting and shadow effects. For each pair of results in row (a)-(c), we display the composite image (\textit{left}) and the harmonized image (\textit{right}). In row (d), we omit the composition for better visibility. Full visualization is provided in the Appendix.}
\label{fig:real_world}
\end{figure*}
  \begin{table*}[ht]
\footnotesize
\centering
\vspace{-2.5mm}
\caption{Ablation to verify the effects of lighting conditioning (`Cond'), alignment (`Align'), and finetuning (`Finetune').}
\vspace{-2.5mm}
\label{table:ablation}
\setlength{\tabcolsep}{2.5pt}
\begin{tabular}{cccccccccccc}
\toprule
\multirow{2}{*}{Model} & \multirow{2}{*}{Cond} & \multirow{2}{*}{Align} & \multirow{2}{*}{Finetune} & \multicolumn{4}{c}{Light stage test set} & \multicolumn{4}{c}{Natural image test set} \\
\cmidrule(lr){5-8}\cmidrule(lr){9-12}
& & & & MSE $\downarrow$ & PSNR $\uparrow$ & SSIM $\uparrow$ & LPIPS $\downarrow$ & MSE $\downarrow$ & PSNR $\uparrow$ & SSIM $\uparrow$ & LPIPS $\downarrow$ \\
\midrule
0 & - & - & - & 0.018 (0.015) & 18.768 (3.531) & 0.815 (0.085) & 0.188 (0.070) & 0.012 (0.008) & 20.158 (3.024) & 0.866 (0.045) & 0.103 (0.028) \\
1 & Bg & - & - & 0.014 (0.012) & 19.748 (3.161) & 0.835 (0.084) & 0.168 (0.063) & 0.013 (0.008) & 19.787 (2.684) & 0.864 (0.042) & 0.106 (0.028) \\
2 & Env & - & - & 0.009 (0.009) & 21.626 (3.162) & 0.866 (0.077) & 0.148 (0.056) & 0.013 (0.008) & 19.824 (2.634) & 0.863 (0.045) & 0.105 (0.028) \\
3 & Bg & $\checkmark$ & - & 0.012 (0.009) & 20.439 (2.987) & 0.842 (0.077) & 0.163 (0.059) & 0.012 (0.007) & 20.006 (2.586) & 0.866 (0.044) & 0.106 (0.028) \\
4 & Bg & $\checkmark$ & $\checkmark$ & 0.012 (0.010) & 20.527 (3.136) & 0.848 (0.076) & 0.159 (0.058) & 0.005 (0.004) & 23.562 (2.830) & 0.913 (0.035) & 0.097 (0.044) \\
\bottomrule
\end{tabular}
\end{table*}

  \begin{figure*}[t]
\centering
\vspace{1mm}
\begin{overpic}[width=0.99\textwidth]{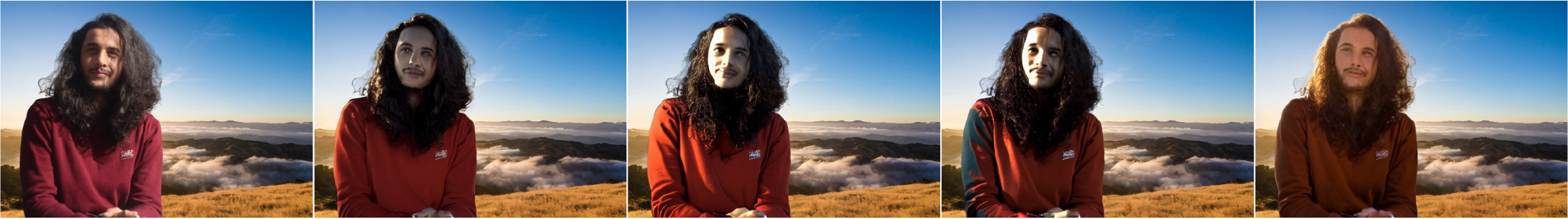}
    \put(6,14.2){\small Composite}
    \put(26,14.2){\small Model 1}
    \put(46,14.2){\small Model 2}
    \put(66,14.2){\small Model 3}
    \put(86,14.2){\small Model 4}
    \end{overpic}
\caption{Example testing results from our ablation. Model 1 to Model 4 correspond to the configurations in Table~\ref{table:ablation}. }
\label{fig:test-ablation}
\end{figure*}
 \begin{figure}
\centering
\begin{overpic}[width=0.45\textwidth]{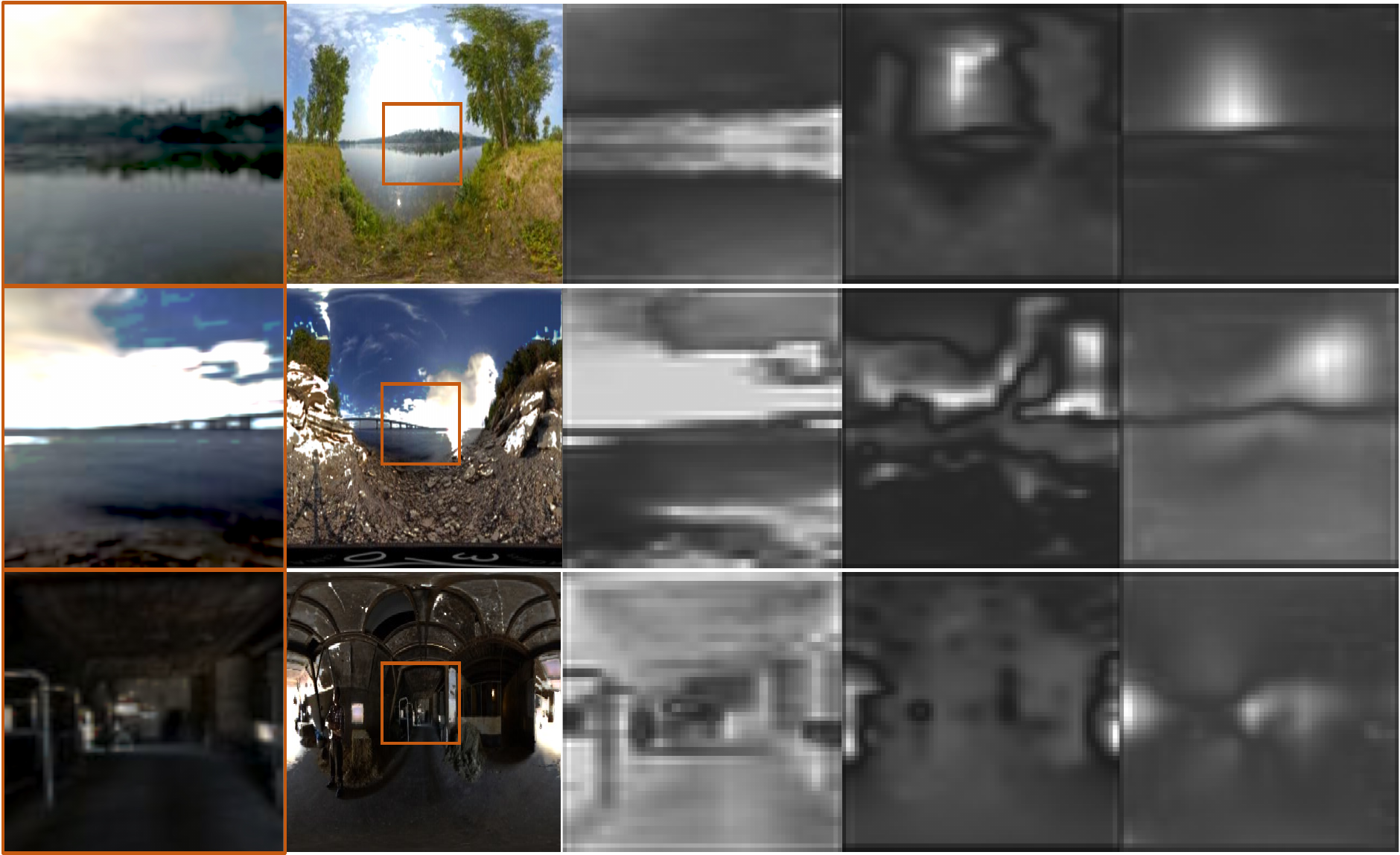}
    \put(8,62){\small Bg}
    \put(27,62){\small Env}
    \put(45,62){\small $\|f_{\text{bg}}\|_2$}
    \put(65,62){\small $\|f_{\text{env}}\|_2$}
    \put(82,62){\small  $\|f_{\text{bg}\rightarrow\text{env}}\|_2$}
    \end{overpic}
\caption{The $L_2$ norm of learned lighting representations. The aligned background-derived feature on the right matches the panorama much closer, indicating a better lighting representation.} 
\label{fig:feature}
\end{figure}

\subsection{Arbitrary Portrait Background Replacement}
We further test our model on various in-the-wild portraits by compositing them onto arbitrary natural backgrounds. 

\noindent\textbf{Lighting Plausibility:} We start by evaluating the lighting plausibility and visual fidelity by replacing the backgrounds with strong lighting indications, like sunlight. Example results are shown in Fig.~\ref{fig:flip}. As can be seen, the lighting tone and direction in the harmonized foreground matched the background effectively. We further vertically flip the backgrounds on the right side, and observe a corresponding change in the lighting effects as expected. 

\noindent\textbf{Shadow Plausibility:}
We then examine how our model handles shadows. 
 Fig.~\ref{fig:shadow-remove} depicts the test cases on input subjects with prominent shadows. When compositing and harmonizing them onto background images with more ambient lighting, our model is able to remove the strong shadow and estimate a shadow-free output, while adapting to the background light. 
Conversely, in Fig.~\ref{fig:shadow-synth}, under backgrounds in daytime scenes with potential overhead sunlight, our model generates visually plausible self-occlusion shadows.

\noindent{\textbf{Creative Background Replacement:}}
Our method allows for creative background replacement for portrait images. In Fig.~\ref{fig:spatiotemporal} (\textit{right}), we create a sequence of background crops from one panorama image with a sliding window, so that the major light source consistently changes from left to right. We observe visually consistent lighting changes on the foreground. Similarly, Fig.~\ref{fig:spatiotemporal} (\textit{left}) shows the harmonization results by placing a subject into different timelapse video frames at different timepoints, resulting in a series of portrait images that effectively mimicked a timelapse effect.\\
\noindent\textbf{Reference-based Harmonization:}
 Our pipeline also extends to reference-based harmonization, where users can integrate their portrait photos with scenes from a chosen reference portrait image. This is achieved by first inpainting the reference portrait to generate a background image, onto which the desired foreground is composited. Our method  ensures that the final output matches the tone and lighting of the original scene. Fig.~\ref{fig:reference} showcases examples where our results effectively align with the tone and lighting of the reference image, displayed on the top left.

\subsection{Ablation}
\label{sec:ablation}
As our proposed method involves multiple stages, we conducted an ablation study to isolate individual and collective effects. We define the base \texttt{Model\#0} as a baseline diffusion model without lighting conditioning. \texttt{Model\#1} and \texttt{Model\#2} incorporate lighting conditioning via background and environment map, respectively. \texttt{Model\#3} further introduces the alignment module and \texttt{Model\#4} is our final model that includes the finetuning. 
 Table~\ref{table:ablation} illustrates the quantitative performance of each model configuration on both light stage and natural image test sets. We include more visual comparisons in the appendix.

\noindent \textbf{Lighting-conditioning:}
Upon integrating lighting conditioning into \texttt{Model\#0}, we observe notable improvements across all metrics on light stage data with both background conditioned \texttt{Model\#1} and environmental conditioned \texttt{Model\#2}.
\texttt{Model\#2} further outperforms \texttt{Model\#1}, verifying our assumption that an environmental map facilitates a more accurate encoding of lighting.

\noindent\textbf{Lighting Representation Alignment:}
In \texttt{Model\#3}, we introduce the embedding alignment. The improvements from \texttt{Model\#1} to \texttt{Model\#3} validated the benefits of utilizing an additional adaptation step to extract robust lighting representation from a single background image. As shown in Fig.~\ref{fig:feature}, the right column representing the aligned feature norms, shows a visual convergence towards the features extracted from the environment map.

\noindent\textbf{Finetuning for Photorealism:}
 \texttt{Model\#4} is a finetuned version of \texttt{Model\#3} with the newly synthesized data. We observe significant improvements in particular on the natural image test set, verifying our assumption that the finetuning boosts the photorealism on natural images. As can be also seen in examples in Fig.~\ref{fig:test-ablation}, while \texttt{Model\#2} and \texttt{Model\#3} estimate plausible lighting direction, the lighting effects are much less realistic than the finetuned~\texttt{Model\#4}.

\section{Conclusion}
We present Relightful Harmonization, a novel lighting-aware diffusion model to blend advanced lighting effects into foreground portraits when compositing onto diverse background images.  
Limitations exist including a resolution cap of 512x512, potentially affecting facial detail, especially in smaller faces. Also, subtle variations may occur in the subject's clothing and skin tones. Detailed analyses and additional failure cases are included in the appendix.

\section{Acknowledgement} We are grateful for Yannick Hold-Geoffroy who provides the panorama environment maps and set up the rendering scripts. We thank Chaowei Company for the support of light stage dataset. We thank David Futschik for running the testing on Total Relighting. We thank Scott Cohen for the inspiration for our project name.
\newpage
{
    \small
    \bibliographystyle{ieeenat_fullname}
    \bibliography{egbib}
}
\clearpage
\setcounter{page}{1}
\maketitlesupplementary

\section{Additional Results}
\label{sec:additional_results}
We note that the all testing input portrait images shown in our paper are sampled from Unsplash or Adobe Stock.

\begin{figure*}[t]
\centering
\vspace{1mm}
\begin{overpic}[width=0.99\textwidth]{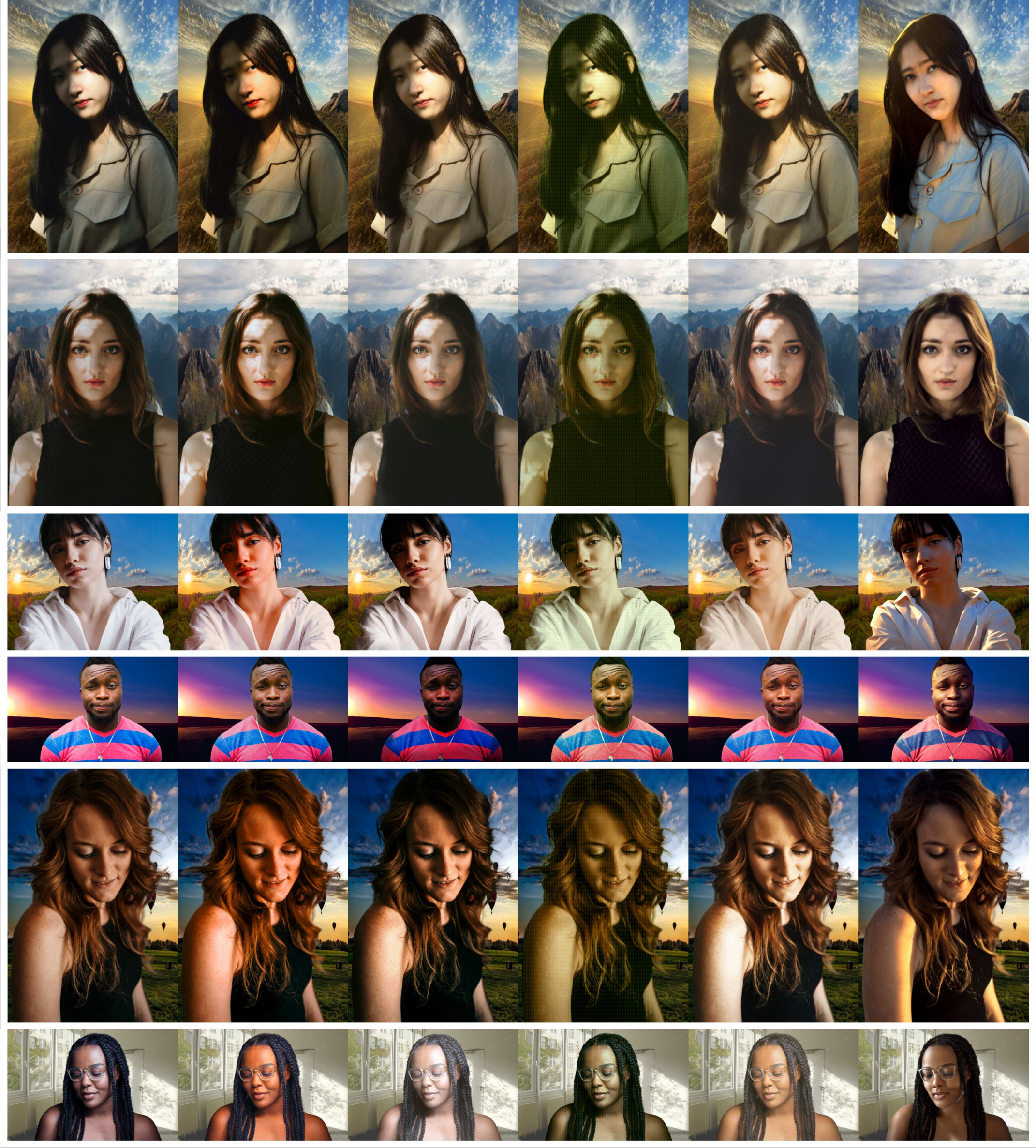}
    \put( 4,101){\small Composite}
    \put(19,101){\small Harmonizer}
    \put(36,101){\small PIH}
    \put(51,101){\small PCT}
    \put(65,101){\small INR}
    \put(80,101){\small Ours}
    \end{overpic}
\caption{Example comparison results on the real world test set to supplement Fig.~\ref{fig:benchmark-test} in the main paper.}
\label{suppl_fig:test-benchmark}
\end{figure*}

\begin{figure*}[t]
\centering
\vspace{1mm}
\begin{overpic}[width=0.99\textwidth]{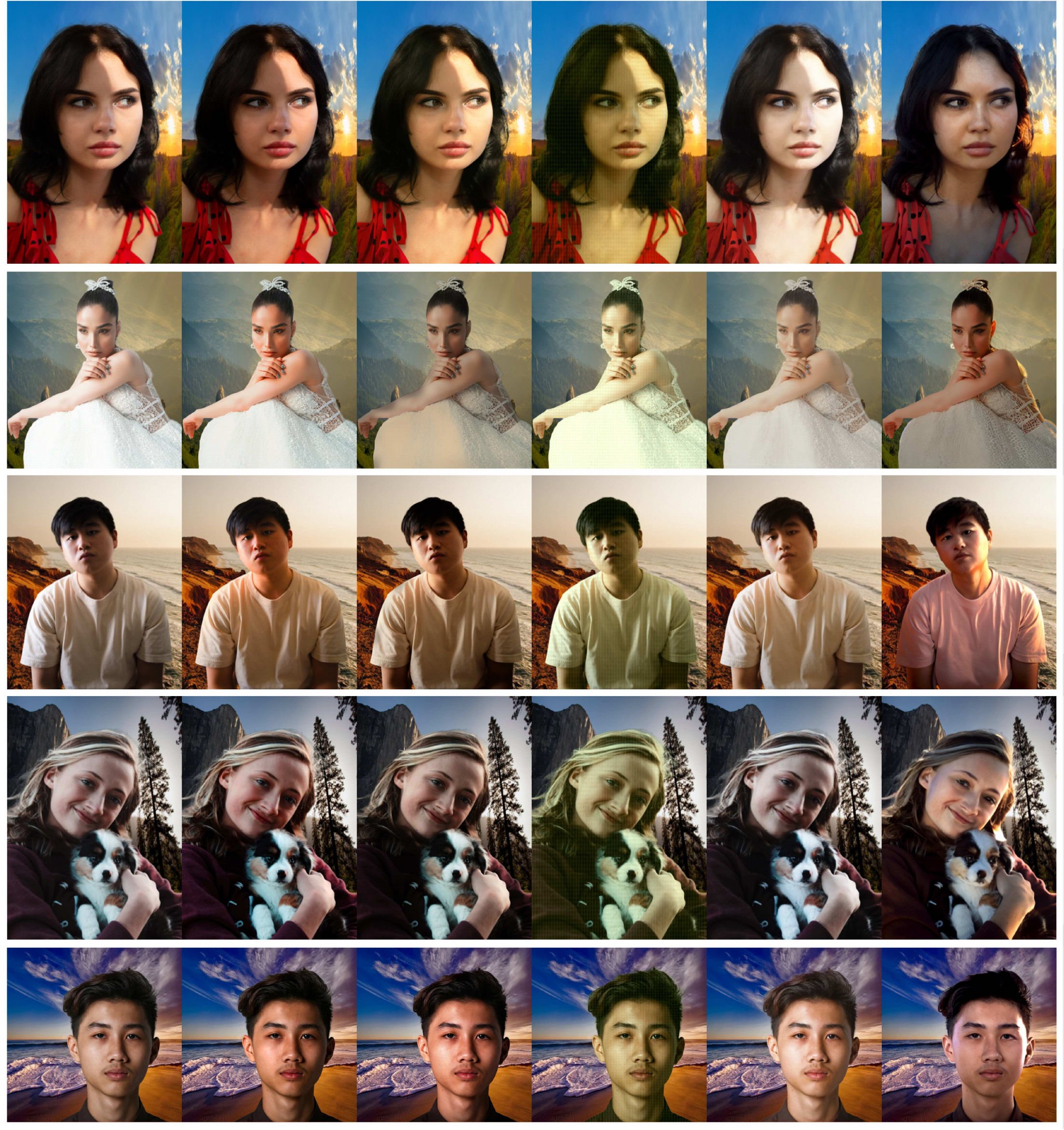}
    \put( 4,101){\small Composite}
    \put(20,101){\small Harmonizer}
    \put(38,101){\small PIH}
    \put(53,101){\small PCT}
    \put(68,101){\small INR}
    \put(84,101){\small Ours}
    \end{overpic}
\caption{Example comparison results on the real world test set to supplement Fig.~\ref{fig:benchmark-test} in the main paper.}
\label{suppl_fig:test2-benchmark}
\end{figure*}

\begin{figure*}[t]
\centering
\vspace{1mm}
\begin{overpic}[width=0.99\textwidth]{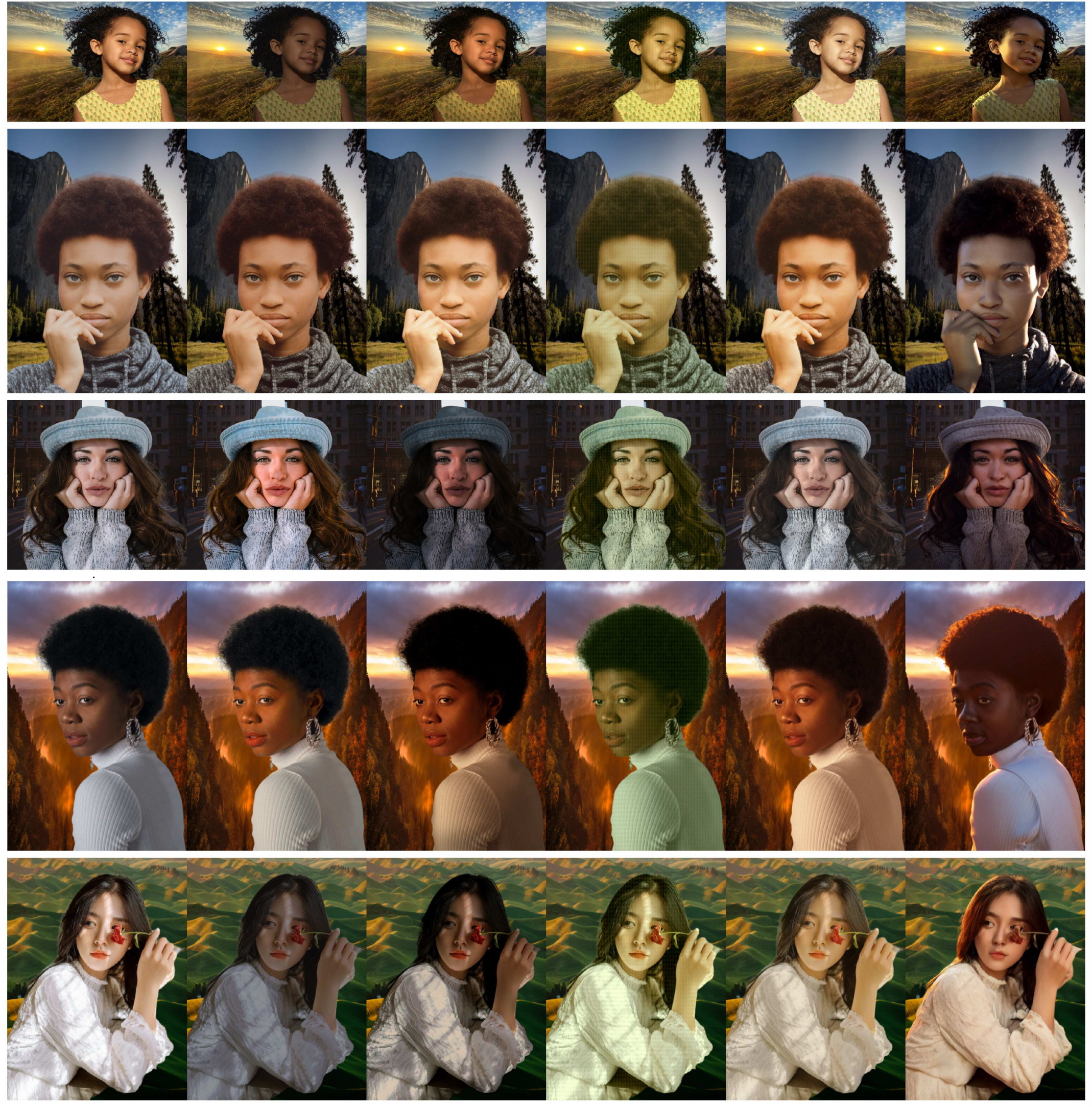}
    \put( 5,101){\small Composite}
    \put(21,101){\small Harmonizer}
    \put(40,101){\small PIH}
    \put(55,101){\small PCT}
    \put(72,101){\small INR}
    \put(88,101){\small Ours}
    \end{overpic}
\caption{Example comparison results on the real world test set to supplement Fig.~\ref{fig:benchmark-test} in the main paper.}
\label{suppl_fig:test3-benchmark}
\end{figure*}

\begin{figure*}[t]
\centering
\vspace{1mm}
\begin{overpic}[width=0.99\textwidth]{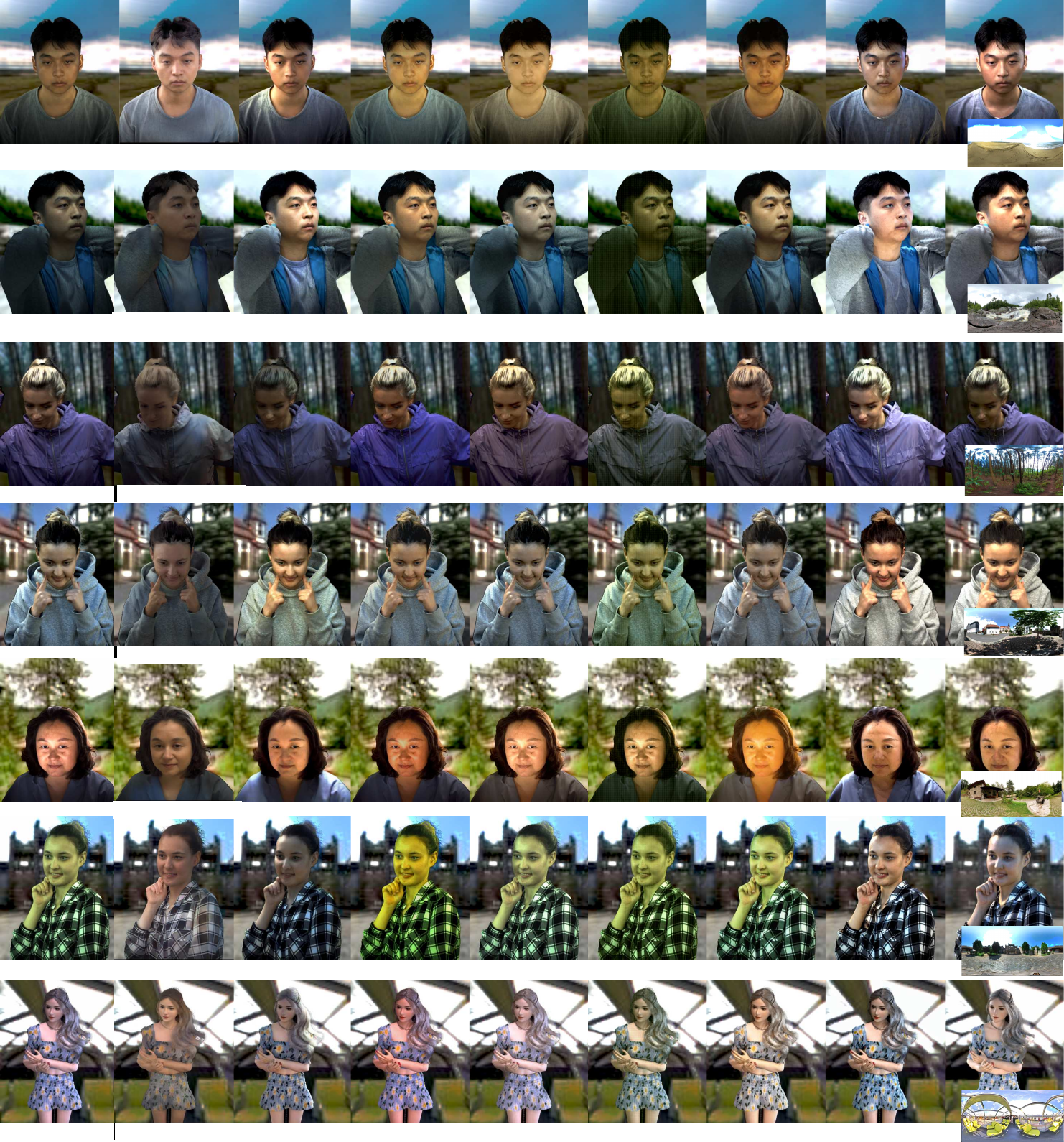}
    \put(1.5,100.5){\small Composite}
    \put(14.5,100.5){\small TR}
    \put(21.5, 100.5){\small Transformer}
    \put(32,100.5){\small Harmonizer}
    \put(45,100.5){\small PIH}
    \put(55,100.5){\small PCT}
    \put(65.5,100.5){\small INR}
    \put(75.5,100.5){\small Ours}
    \put(83.5,100.5){\small Ground Truth}
    \end{overpic}
\caption{Example comparison results on the light stage test set to supplement Fig.~\ref{fig:lightstage} in the main paper. The environment map is shown at the bottom of the ground truth image.}
\label{suppl_fig:lightstage-benchmark}
\end{figure*}

\begin{figure*}[t]
\centering
\vspace{1mm}
\begin{overpic}[width=0.99\textwidth]{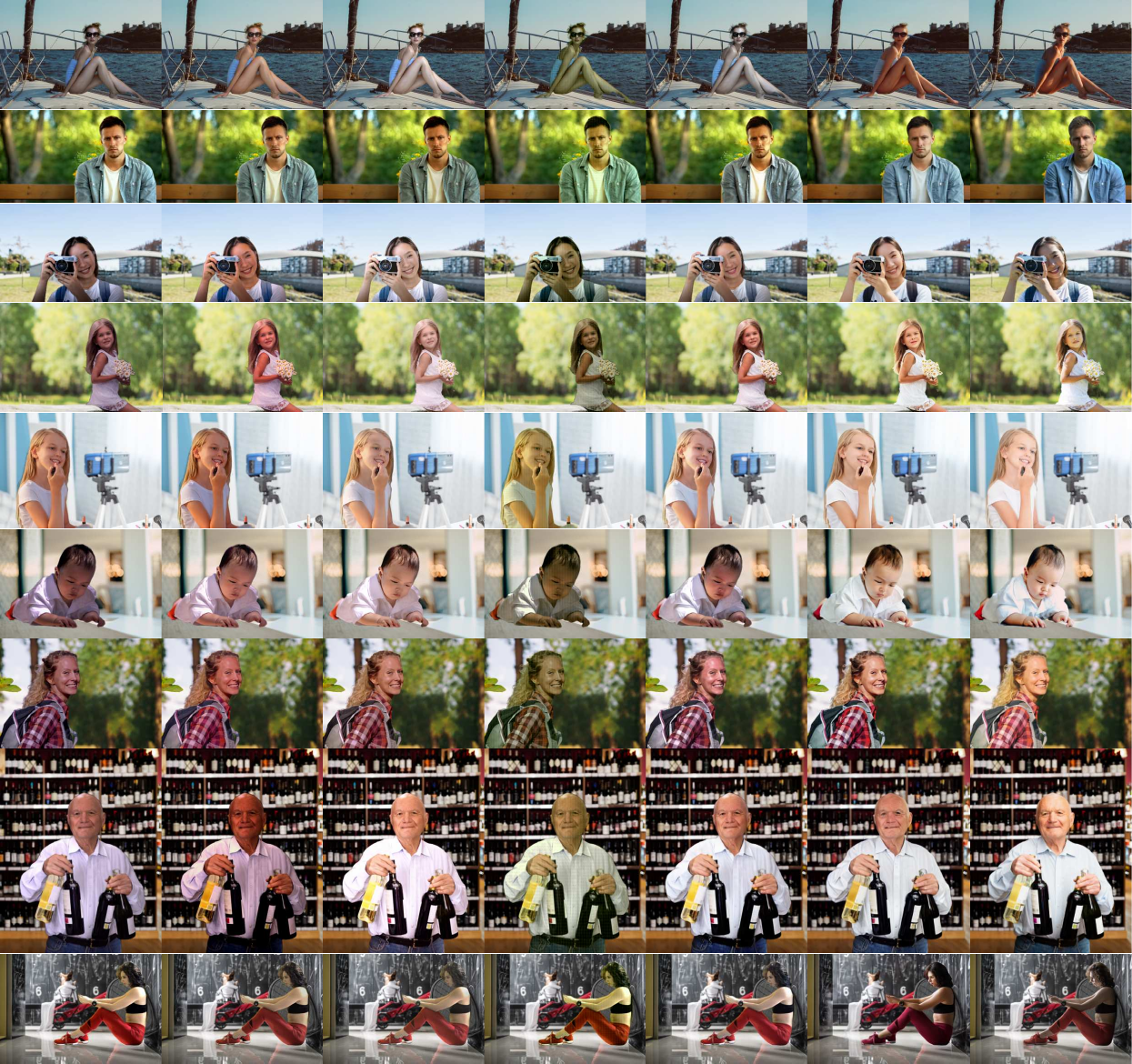}
    \put(3,95){\small Composite}
    \put(17,95){\small Harmonizer}
    \put(34,95){\small PIH}
    \put(48,95){\small PCT}
    \put(63,95){\small INR}
    \put(77,95){\small Ours}
    \put(88,95){\small Ground Truth}
    \end{overpic}
\caption{Example comparison results on the natural image test set to supplement Fig.~\ref{fig:stock} in the main paper.}
\label{suppl_fig:stock-benchmark}
\end{figure*}

\noindent\textbf{Comparison with benchmarks}
To supplement Fig.~\ref{fig:benchmark-test}, we present additional visual comparison with benchmarks on the real world data in Fig.~\ref{suppl_fig:test-benchmark},~\ref{suppl_fig:test2-benchmark} and~\ref{suppl_fig:test3-benchmark}. 
To supplement Fig.~\ref{fig:lightstage} and Fig.~\ref{fig:stock}, we show full benchmark comparison on the light stage test dataset in Fig.~\ref{suppl_fig:lightstage-benchmark}, and on the natural image test set in Fig.~\ref{suppl_fig:stock-benchmark}.

\begin{figure*}[t]
\centering
\vspace{1mm}
\begin{overpic}[width=0.99\textwidth]{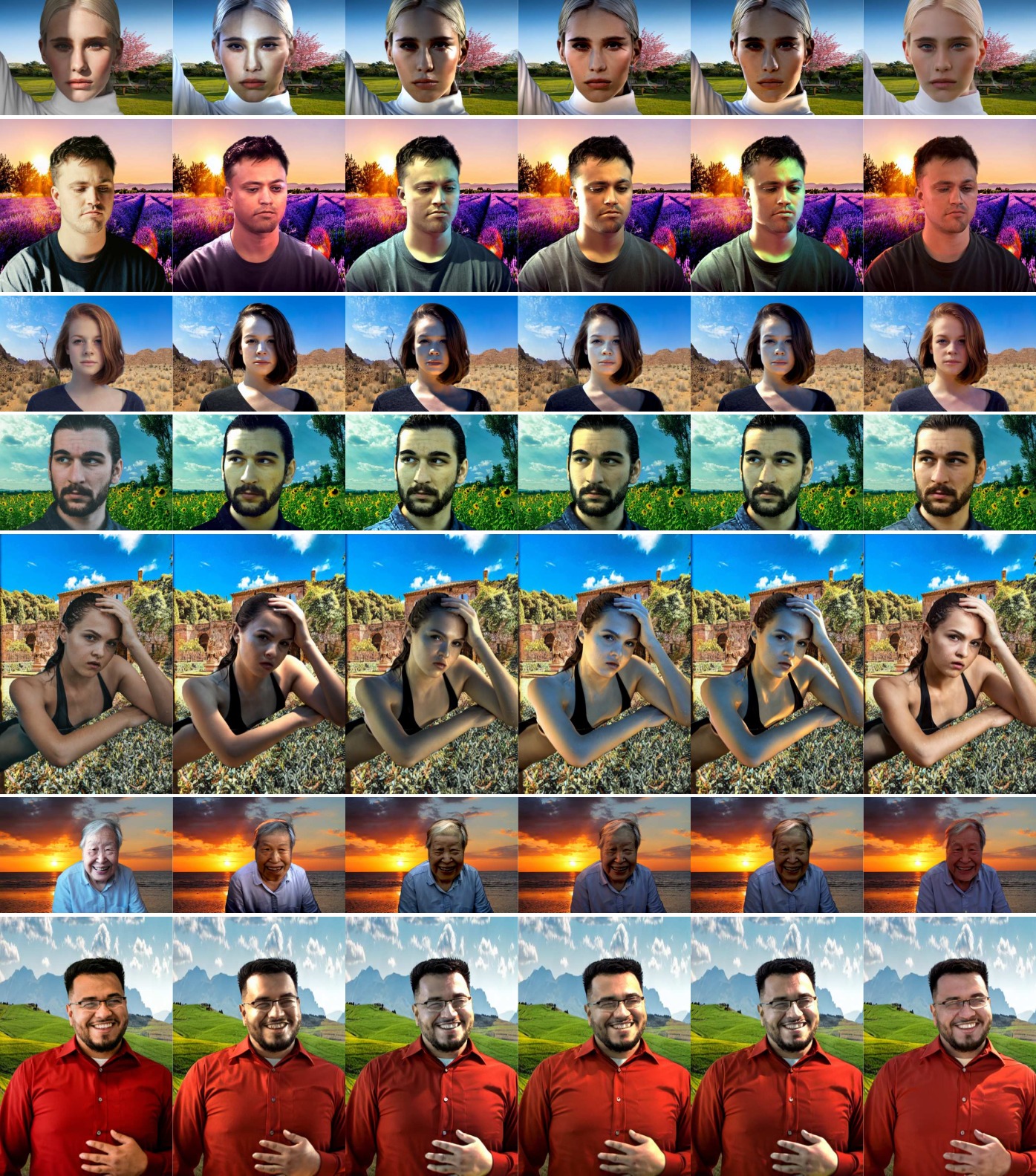}
    \put(3,100.5){\small Composite}
    \put(19,100.5){\small Model 0}
    \put(34,100.5){\small Model 1}
    \put(49,100.5){\small Model 2}
    \put(63,100.5){\small Model 3}
    \put(78,100.5){\small Model 4}
    \end{overpic}
\caption{Example testing results from our ablation on the real image test set. Model 0 to Model 4 correspond to the configurations in Table~\ref{table:ablation}. Our final model (Model 4) presents the best visual quality while maintaining plausible lighting effects. }
\label{suppl_fig:real-ablation}
\end{figure*}

\begin{figure*}[t]
\centering
\vspace{1mm}
\begin{overpic}[width=0.99\textwidth]{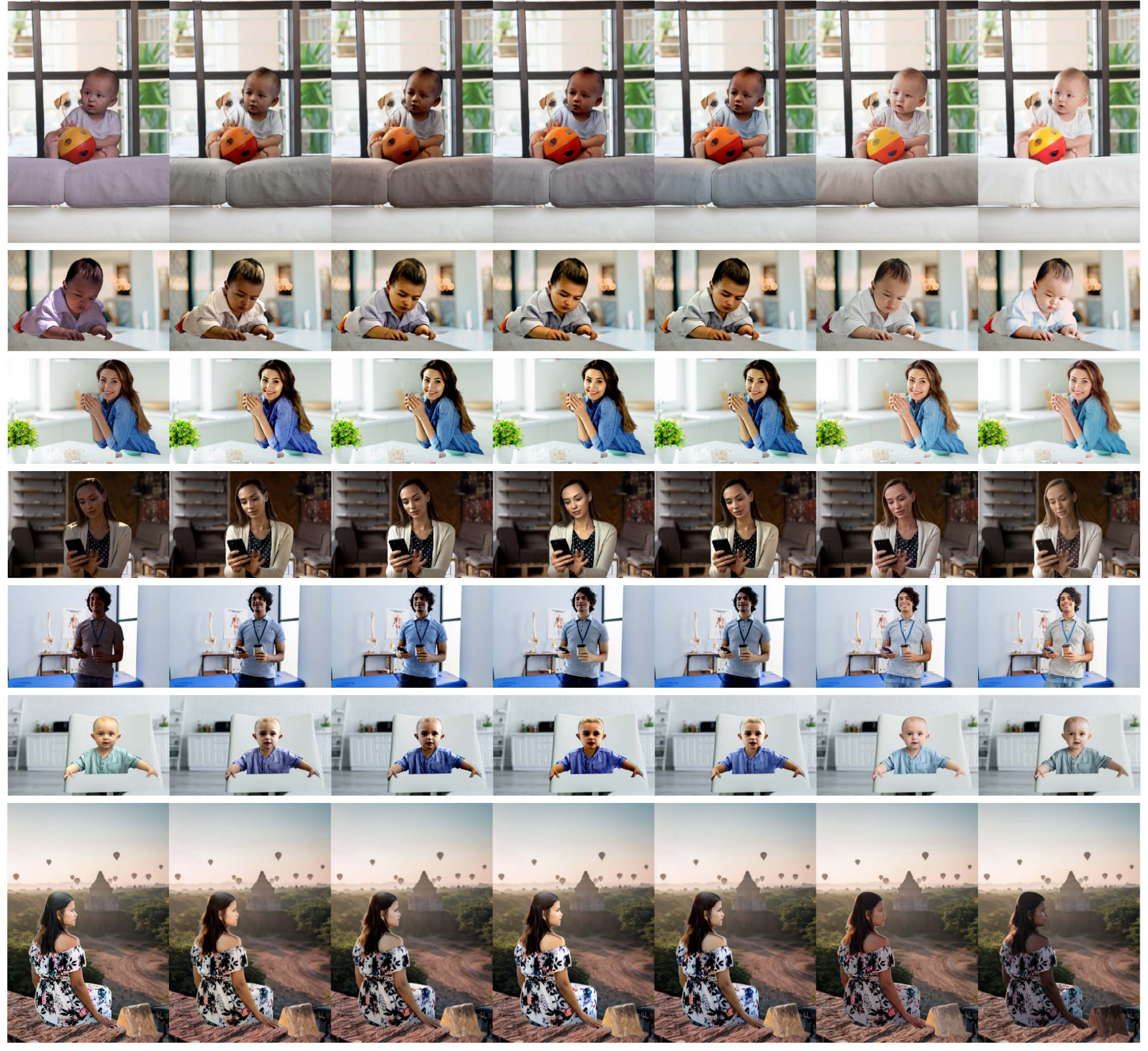}
    \put(4,92){\small Composite}
    \put(19,92){\small Model 0}
    \put(33,92){\small Model 1}
    \put(47,92){\small Model 2}
    \put(61,92){\small Model 3}
    \put(75,92){\small Model 4}
    \put(87,92){\small Ground Truth}
    \end{overpic}
\caption{Example testing results from our ablation on the natural image test set. Model 0 to Model 4 correspond to the configurations in Table~\ref{table:ablation}. Our final model (Model 4) presents the best visual quality while maintaining plausible lighting effects. }
\label{suppl_fig:stock-ablation}
\end{figure*}

\begin{figure*}[t]
\centering
\vspace{1mm}
\begin{overpic}[width=0.99\textwidth]{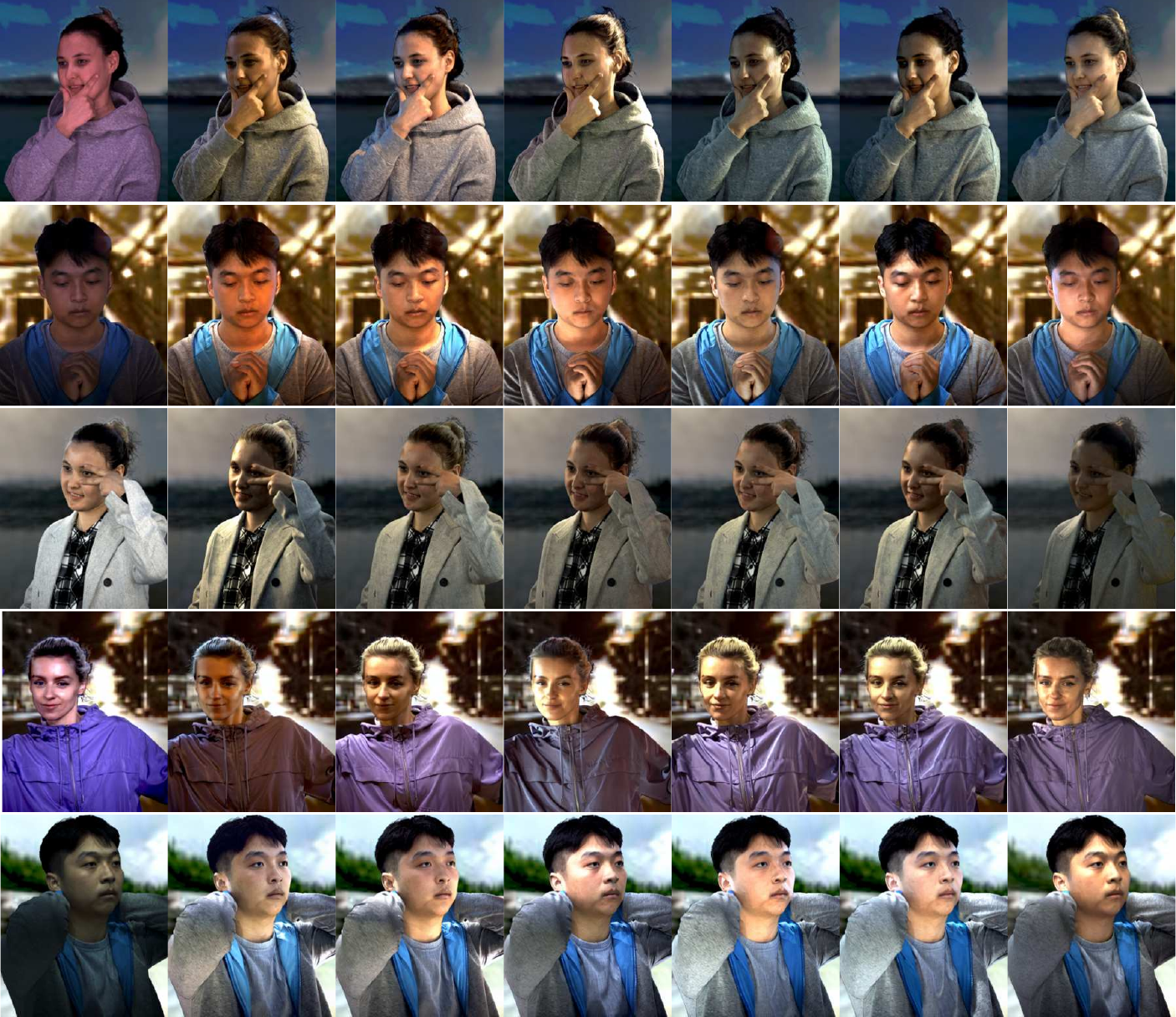}
    \put(3,87){\small Composite}
    \put(19,87){\small Model 0}
    \put(33,87){\small Model 1}
    \put(47,87){\small Model 2}
    \put(62,87){\small Model 3}
    \put(76,87){\small Model 4}
    \put(88,87){\small Ground Truth}
    \end{overpic}
\caption{Example testing results from our ablation on the light stage test set. Model 0 to Model 4 correspond to the configurations in Table~\ref{table:ablation}. Our final model (Model 4) presents the best visual quality while maintaining plausible lighting effects. }
\label{suppl_fig:lightstage-ablation}
\end{figure*}

\begin{figure*}
\centering
\begin{overpic}[width=0.99\textwidth]{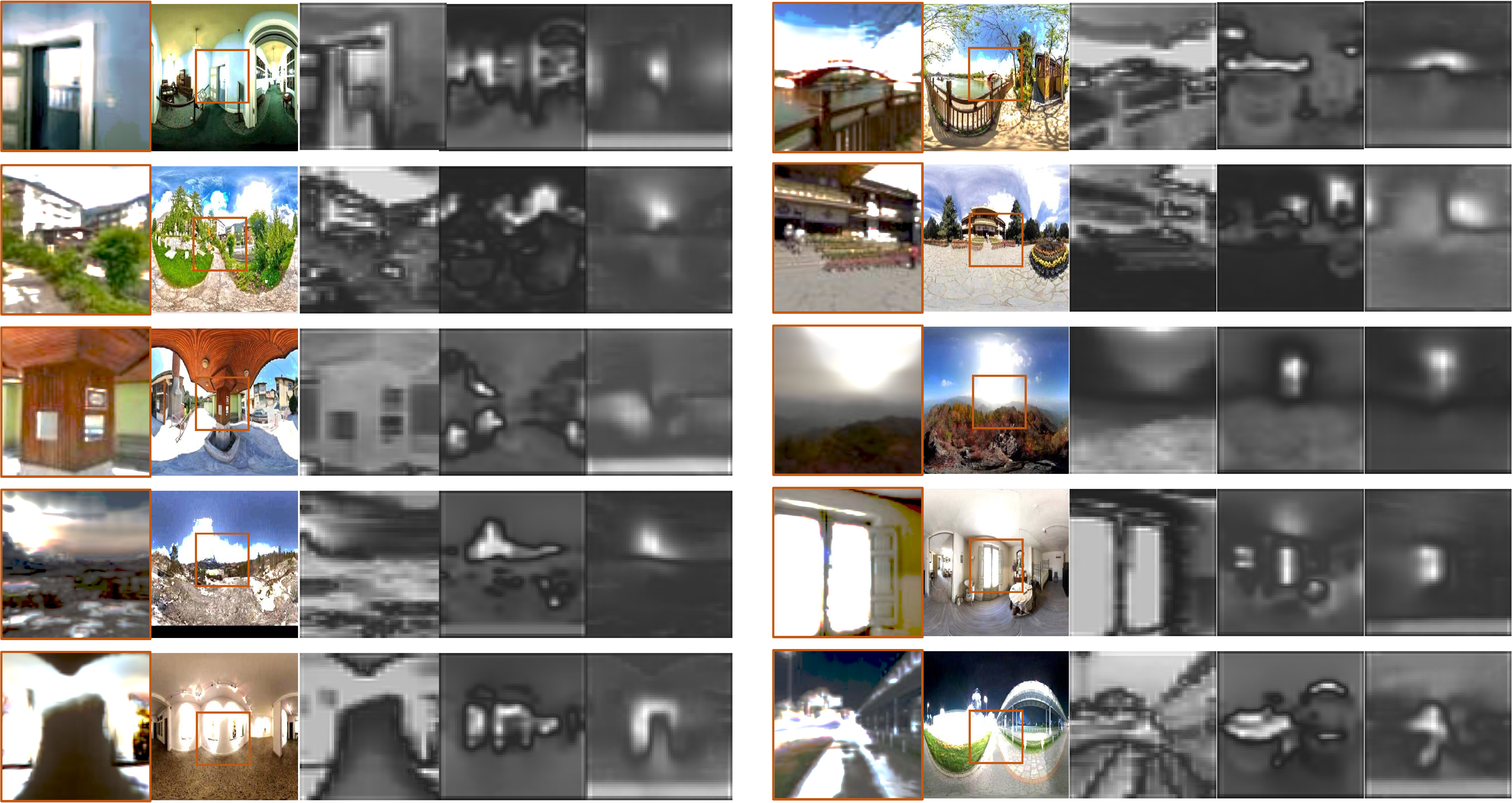}
    \put(4,54){\small Bg}
    \put(13.5,54){\small Env}
    \put(22.5,54){\small $\|f_{\text{bg}}\|_2$}
    \put(31.5,54){\small $\|f_{\text{env}}\|_2$}
    \put(40,54){\small  $\|f_{\text{bg}\rightarrow\text{env}}\|_2$}
    \put(55,54){\small Bg}
    \put(64.5,54){\small Env}
    \put(72.5,54){\small $\|f_{\text{bg}}\|_2$}
    \put(82.5,54){\small $\|f_{\text{env}}\|_2$}
    \put(91,54){\small  $\|f_{\text{bg}\rightarrow\text{env}}\|_2$}
    \end{overpic}
\caption{The $L_2$ norm of learned lighting representations to supplement Fig.~\ref{fig:feature}. The aligned background-derived feature on the right matches the panorama much closer, indicating a better lighting representation.} 
\label{suppl_fig:feature}
\end{figure*}

\noindent\textbf{Ablation}
We present additional visual comparison among our ablation models on the natural image test and light stage test set in Fig.~\ref{suppl_fig:stock-ablation} and  Fig.~\ref{suppl_fig:lightstage-ablation} respectively. 
The ablation comparison on the real test set is shown in Fig.~\ref{suppl_fig:real-ablation} as a supplement to Fig.~\ref{fig:test-ablation}. 
In Fig.~\ref{suppl_fig:feature}, we include additional visualization of the feature norms to illustrate the affects of the alignment module to supplement Fig.~\ref{fig:feature}.

\begin{figure*}[t]
    \centering
    \includegraphics[width=0.99\linewidth]{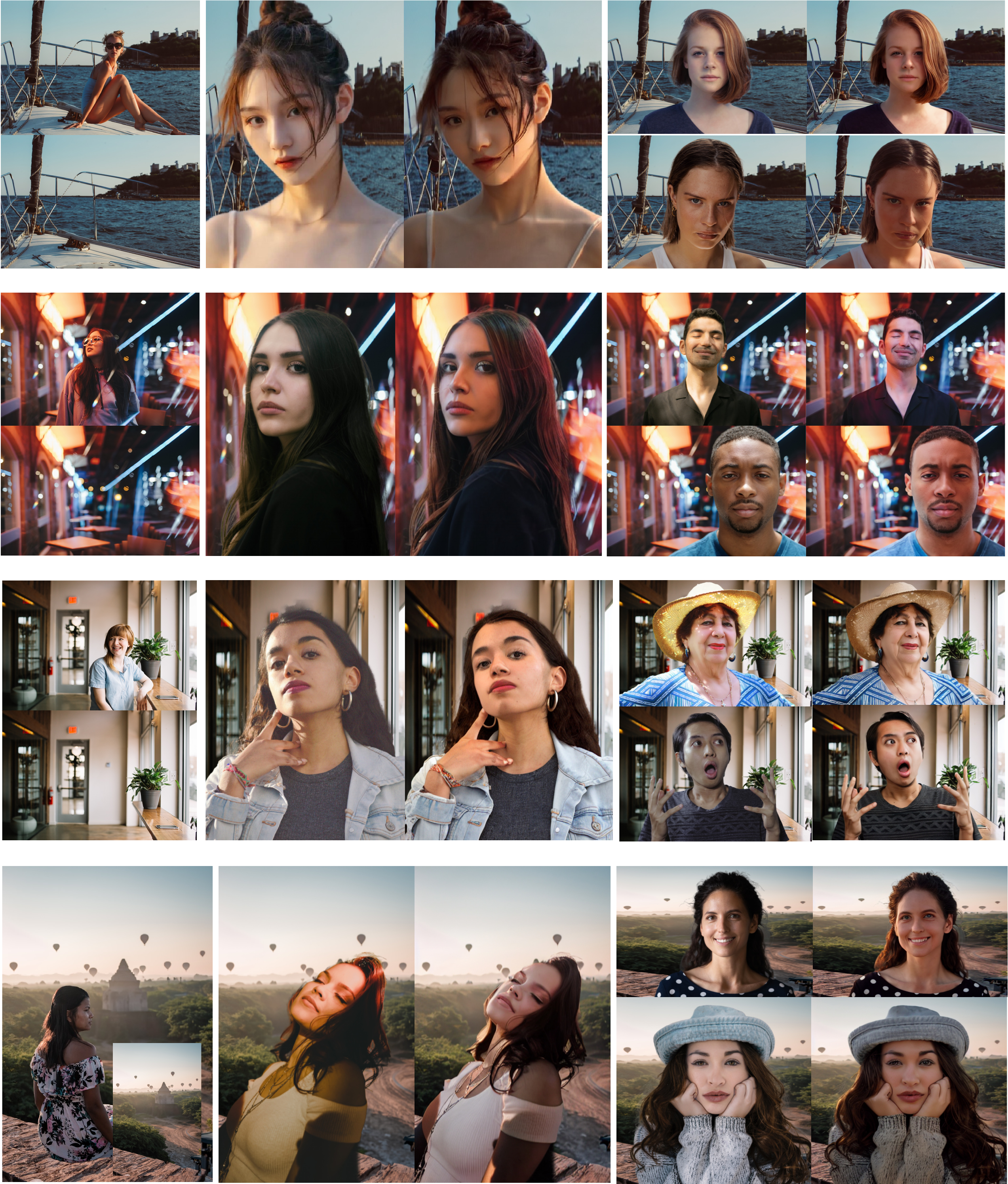}
\vspace{-2mm}
    \caption{Visual results on the reference-based harmonization application to supplement Fig.~\ref{fig:reference}. It allows user images to be blended into scenes from real portraits. This involves removing the subject from the reference image (\textit{upper left}) to create a background (\textit{lower left}) for composition. The harmonized results (\textit{right}) achieve lighting effects closely resembling those in the reference.}
    \label{suppl_fig:reference}
\end{figure*}

\begin{figure*}[t]
    \centering
    \includegraphics[width=0.99\linewidth]{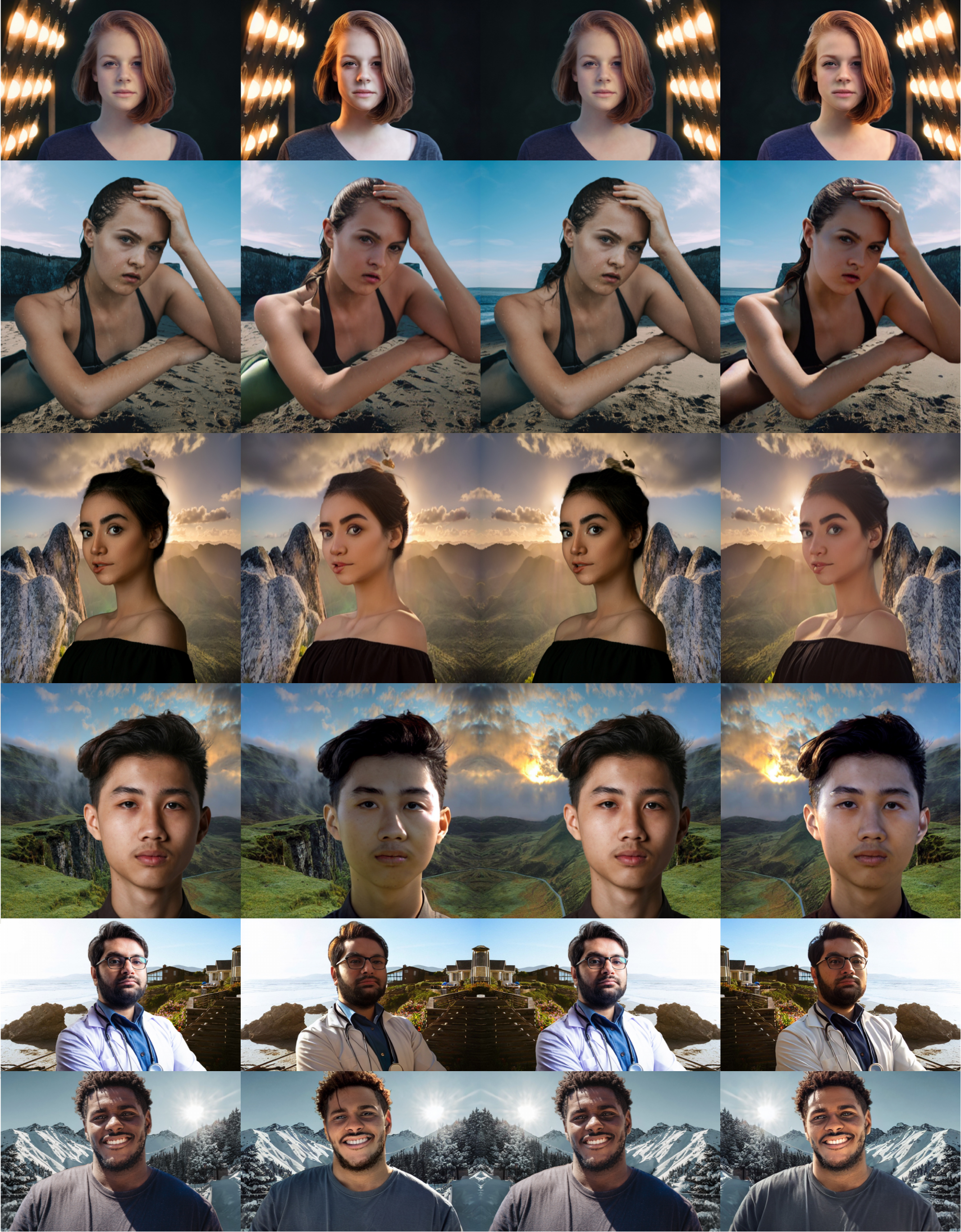}
\vspace{-2mm}
    \caption{Harmonization results when flipping the background, to supplement Fig.~\ref{fig:flip}.}
    \label{suppl_fig:flip}
\end{figure*}

\begin{figure*}
\centering
\vspace{10mm}
\begin{subfigure}[][][t]{0.96\textwidth}
    \includegraphics[width=\textwidth]{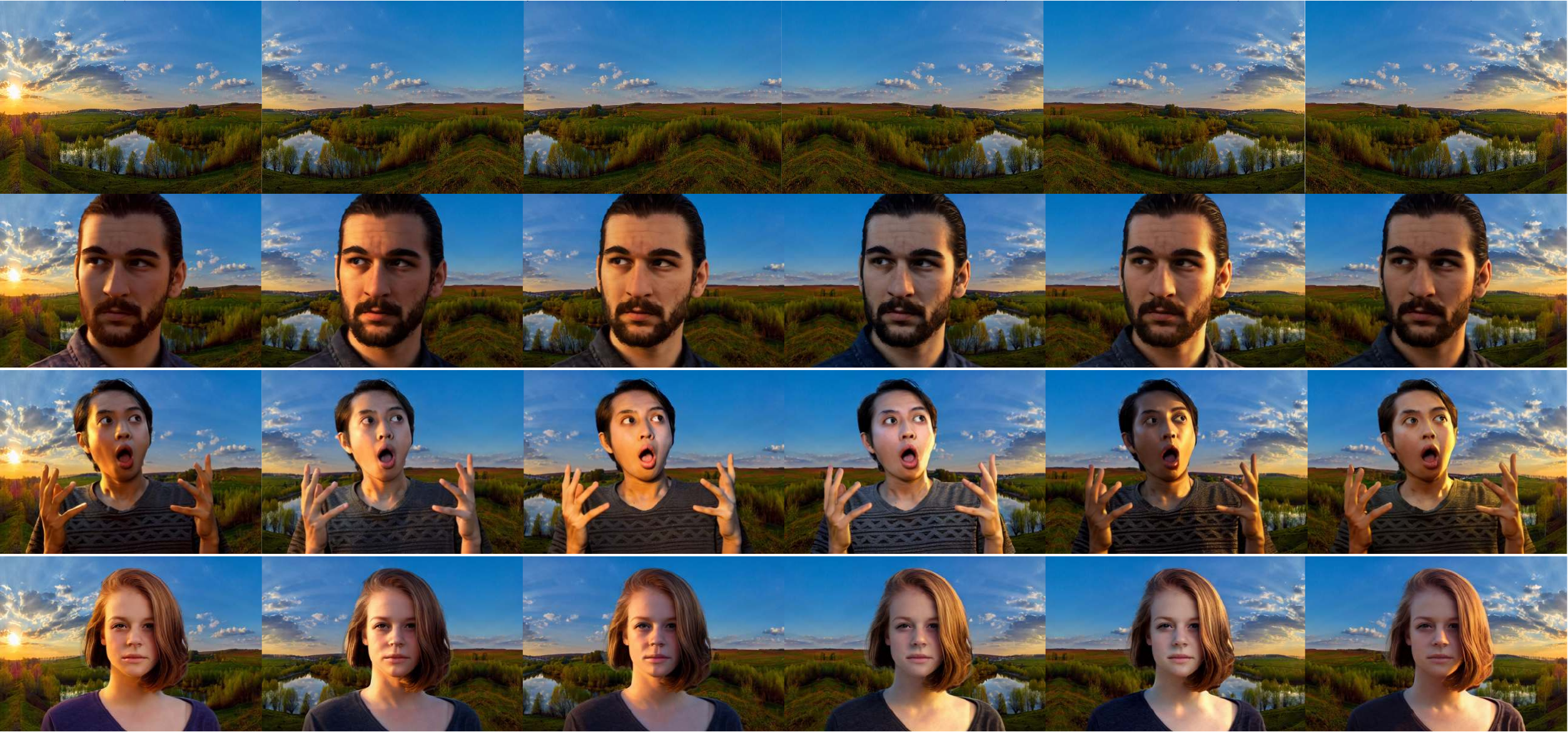}
    \caption{We create spatially changing lighting conditions by cropping background images (\textit{top}) from a panoramic image. Our model produces visually coherent lighting changes on different portrait images.}
    \label{suppl_fig:rotate_bg}
\end{subfigure}
\unskip\
\begin{subfigure}[][][t]{0.97\textwidth}
\vspace{10mm}
    \includegraphics[width=\textwidth]{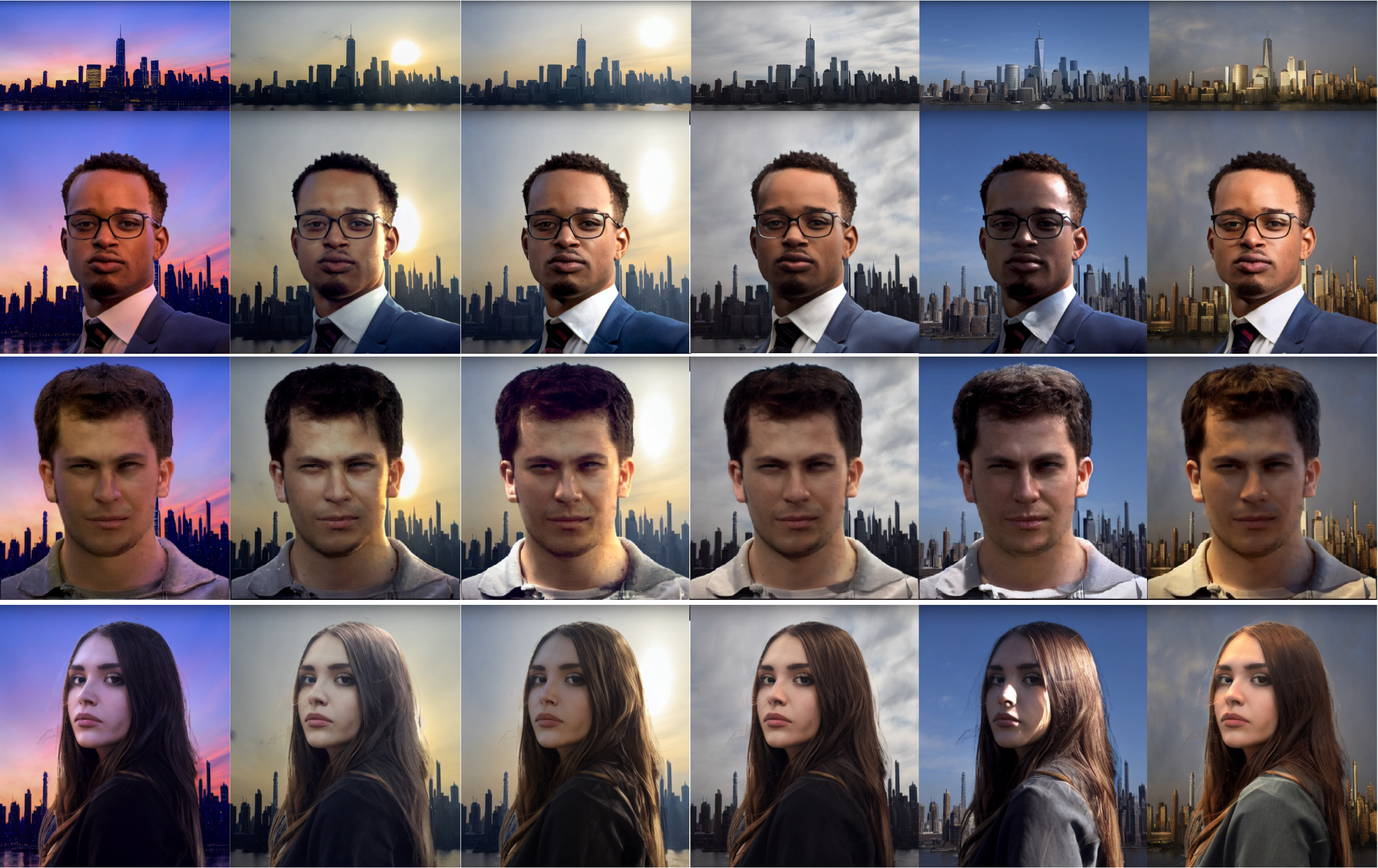}
    \caption{We obtain temporally changing lighting conditions by taking multiple screenshots (\textit{top}) from a timelapse video (\url{https://www.youtube.com/watch?v=CSfri4U9w28}). Our model produces visually reasonable harmonization results.}
    \label{suppl_fig:temporal}
\end{subfigure}

\caption{Harmonization results under the background images where lighting conditions are changing spatially (a) or temporally (b).}
\label{suppl_fig:spatiotemporal}
\end{figure*}

\noindent\textbf{Real world testing results}
Fig.~\ref{suppl_fig:reference} shows reference based harmonization example as in Fig.~\ref{fig:reference}.
Fig.~\ref{suppl_fig:flip} shows harmonization results when we flip the background image as in Fig~\ref{fig:flip}. 
Fig.~\ref{suppl_fig:spatiotemporal} shows the results under spatially and temporally changing lighting as in Fig.~\ref{fig:spatiotemporal}.

\section{Additional Implementation Details}
\subsection{Network Architecture}

\noindent\textbf{Lighting-conditioned diffusion} is built on the InstructPix2Pix~\cite{InstructPix2Pix} backbone.  The core rationale for selecting the pretrained InstructPix2Pix model \cite{InstructPix2Pix} as the foundation, rather than the stable diffusion model, is on its capability to incorporate an additional input image channel. Therefore, at the beginning of our training, the input and output image will be identital (i.e., no editing on the input image), and it will gradually incorporate the lighting conditioning from the extra lighting representation. We use a dummy editing prompt \textit{`portrait'} during our training and inference.

The lighting conditioning branch architecture follows ControlNet~\cite{ControlNet}, where an encoder structure identical to the diffusion UNet backbone is applied and the intermediate feature maps are added to the UNet encoder at respective resolutions. The lighting representation is extracted from a 4-layer CNN. We train our model with the input resolution of $512\times512$ (for both input image and the background image), and the lighting representation is a tensor with shape $64 \times 64\times 320$. We empirically found that training with a higher resolution (e.g., $768\times768$) led to better identity preservation, but performed worse in terms of the relighting. We speculate that this is related to the stable diffusion pretraining, which is on $512\times 512$ resolution.

\noindent\textbf{The Alignment Network} is an encoder-decoder architecture built with Residual blocks. The encoder is composed of three sequential residual blocks. Each of these blocks is coupled with a subsequent downsampling layer. The decoder is symmetrical to the encoder, with three residual blocks, and each of them followed by an upsampling layer. The input and output dimensions of the alignment network are consistent, maintaining a shape of $64 \times 64 \times 320$.

\noindent\textbf{Ablation Models Specifics} 
\texttt{Model\#0} is a baseline diffusion model without lighting conditioning and its implementation follows InstructPix2Pix~\cite{InstructPix2Pix} with the text prompt fixed as `Portrait'. \texttt{Model\#1} takes the background image as the conditional input, which is resized to $512\times 512$. \texttt{Model\#2} shares the same architecture as \texttt{Model\#1} but replaces the conditional input to the LDR environment map. 
\texttt{Model\#3} introduces the alignment module after the conditional branch from \texttt{
Model\#1}. The Unet backbone from \texttt{Model\#2} is used as diagramed in Fig.~\ref{fig:overview}.
\texttt{Model\#4} finetunes on \texttt{Model\#3} with the synthetic data. 

\subsection{Transformer relighting model}
To train a relighting baseline on our light stage dataset, we built a transformer based encoder-decoder network. The network input is a concatenated input image, foreground mask, and the parsing mask, which is divided into patches of $4\times4$. A hierarchical Transformer encoder is applied to obtain multi-level features at $\{\frac{1}{4},\frac{1}{8}, \frac{1}{16},\frac{1}{32}\}$ of the original resolution. A decoder with transpose convolution is then followed to get the final result with the same resolution as the input. The target LDR environment map is concatenated at the bottleneck latent space in a similar manner as~\cite{sun2019single}.

\section{Failure case and analysis}
\begin{figure*}
\centering
\vspace{10mm}
\begin{subfigure}[][][t]{0.96\textwidth}
    \includegraphics[width=\textwidth]{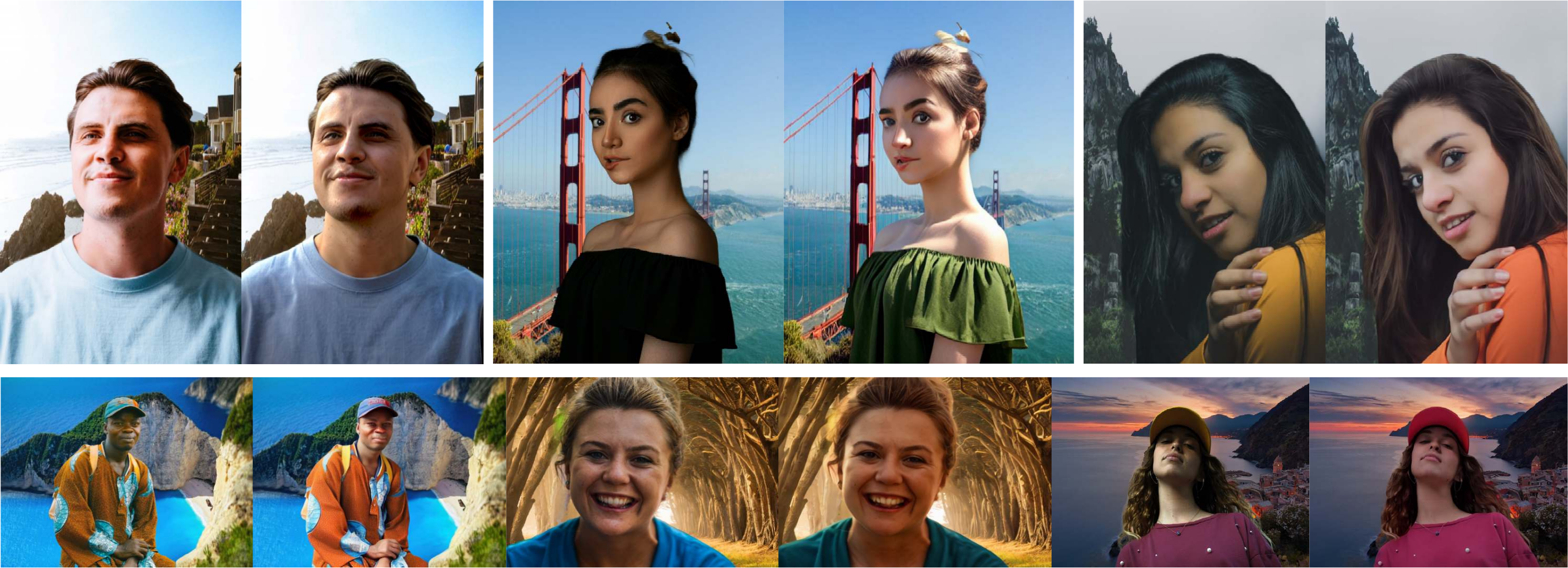}
    \caption{In some examples, our model modified the color of the subject clothes due to its harmonization nature.}
    \label{suppl_fig:fail_cloth_color}
\end{subfigure}
\unskip\
\begin{subfigure}[][][t]{0.97\textwidth}
\vspace{10mm}
    \includegraphics[width=\textwidth]{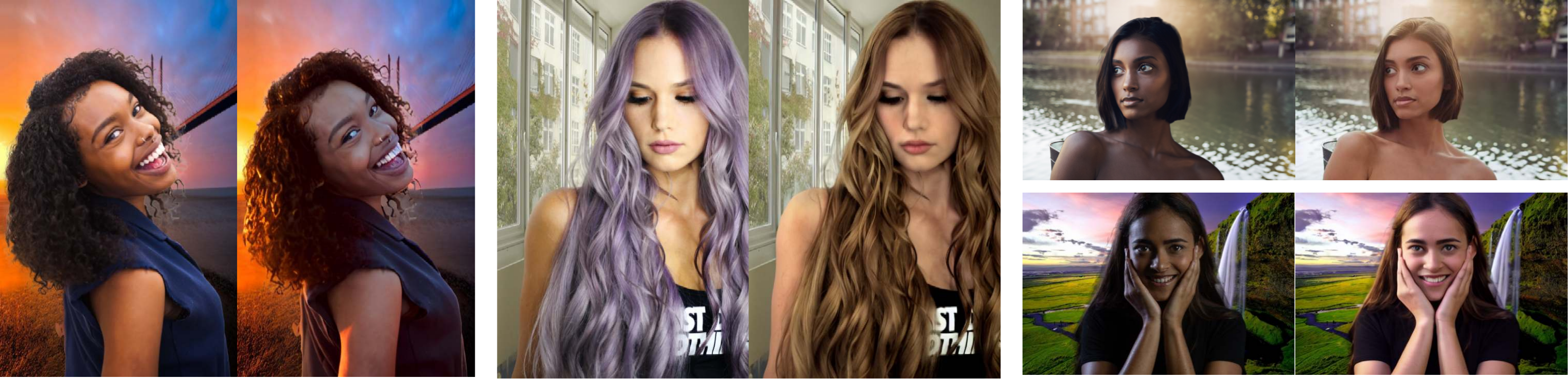}
    \caption{In some examples, our model may not fully preserve the subject identity, such as the hair color and the skin tone, especially when the input skin tone is ambiguous (right two examples).}
    \label{suppl_fig:fail_identity}
\end{subfigure}

\unskip\ 
\begin{subfigure}[][][t]{0.97\textwidth}
\vspace{10mm}
    \includegraphics[width=\textwidth]{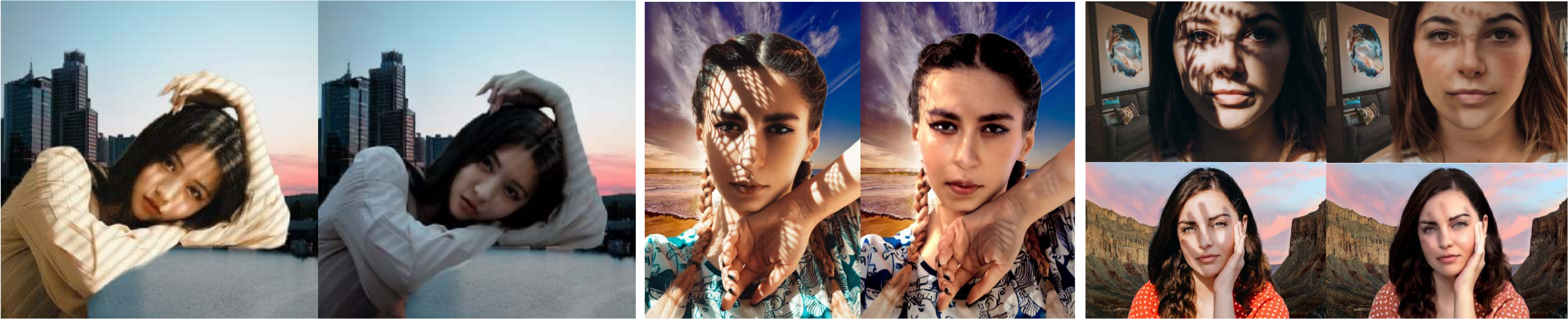}
    \caption{In portraits with strong casted shadows, our model may fail to completely remove them.}
    \label{suppl_fig:fail_shadow}
\end{subfigure}

\caption{Failure cases}
\label{suppl_fig:failure}
\end{figure*}

We illustrate several example failure cases in Fig.\ref{suppl_fig:failure}. In our training approach, since we do not impose constraints on the subject's identity, there are instances where the model struggles to retain identity-specific details. For instance, as shown in Fig.\ref{suppl_fig:fail_cloth_color}, the color of the subject’s clothing is inaccurately altered during the color harmonization process. Similarly, in Fig.~\ref{suppl_fig:fail_identity}, there is a notable change in hair color (\textit{middle}). Furthermore, in scenarios where the input skin tone is not clearly indicated (\textit{right}), our model occasionally produces ambiguous results in skin tone modification. Additionally, our method does not incorporate intermediate steps like albedo estimation, which can be crucial in handling complex lighting conditions. As a result, in inputs with pronounced cast shadows, our model sometimes fails to eliminate these shadows effectively.
\newpage


\end{document}